\documentclass[]{fairmeta}

\usepackage{times}
\usepackage{latexsym}
\usepackage[T1]{fontenc}
\usepackage[utf8]{inputenc}
\usepackage{microtype}
\usepackage{inconsolata}
\usepackage{graphicx}
\usepackage{algorithm,algorithmic}
\usepackage{amsmath}
\usepackage{cuted}
\usepackage[dvipsnames]{xcolor}
\usepackage{booktabs}
\usepackage{multirow}
\usepackage[normalem]{ulem}
\useunder{\uline}{\ul}{}
\usepackage{listings}
\usepackage{pythonhighlight}
\usepackage{tcolorbox}
\usepackage{amsfonts}
\usepackage{etoolbox}
\usepackage{makecell}
\usepackage{wrapfig}
\AfterEndEnvironment{strip}{\leavevmode}

\definecolor{query}{HTML}{EBC9BC}
\definecolor{retrievertool}{HTML}{3DADFF}
\definecolor{agent}{HTML}{66D575}
\definecolor{workingmemory}{HTML}{F849C1}
\definecolor{database}{HTML}{FF9E42}
\definecolor{extractor}{HTML}{FFC7C2}

\newcommand{\mdlname}[0]{EGAgent}

\title{Agentic Very Long Video Understanding}

\author[1,2,*]{Aniket Rege}
\author[1]{Arka Sadhu}
\author[1]{Yuliang Li}
\author[1]{Kejie Li}
\author[2]{Ramya Korlakai Vinayak}
\author[1]{Yuning Chai}
\author[2]{Yong Jae Lee}
\author[1]{Hyo Jin Kim}

\affiliation[1]{Reality Labs Research at Meta}
\affiliation[2]{University of Wisconsin-Madison}

\contribution[*]{Work done during internship at Meta \vskip -0.1cm}

\abstract{
The advent of always-on personal AI assistants, enabled by all-day wearable devices such as smart glasses, demands a new level of contextual understanding, one that goes beyond short, isolated events to encompass the continuous, longitudinal stream of egocentric video. 
Achieving this vision requires advances in long-horizon video understanding, where systems must interpret and recall visual and audio information spanning days or even weeks. 
Existing methods, including large language models and retrieval-augmented generation, are constrained by limited context windows and lack the ability to perform compositional, multi-hop reasoning over very long video streams. 
In this work, we address these challenges through \textbf{\mdlname{}}, an enhanced agentic framework centered on entity scene graphs, which represent people, places, objects, and their relationships over time. 
Our system equips a planning agent with tools for structured search and reasoning over these graphs, as well as hybrid visual and audio search capabilities, enabling detailed, cross-modal, and temporally coherent reasoning. 
Experiments on the EgoLifeQA and Video-MME (Long) datasets show that our method achieves state-of-the-art performance on EgoLifeQA (57.5\%) and competitive performance on Video-MME (Long) (74.1\%) for complex longitudinal video understanding tasks.
}
\correspondence{Aniket Rege at <\email{aniketr@cs.wisc.edu}>, Hyo Jin Kim at <\email{kimhyojin@meta.com}>}

\metadata[Code]{\url{https://github.com/facebookresearch/egagent}}

\begin{document}

\maketitle
\vspace{1mm}
\begin{center}
\captionsetup{type=figure}
\includegraphics[width=0.88\linewidth]{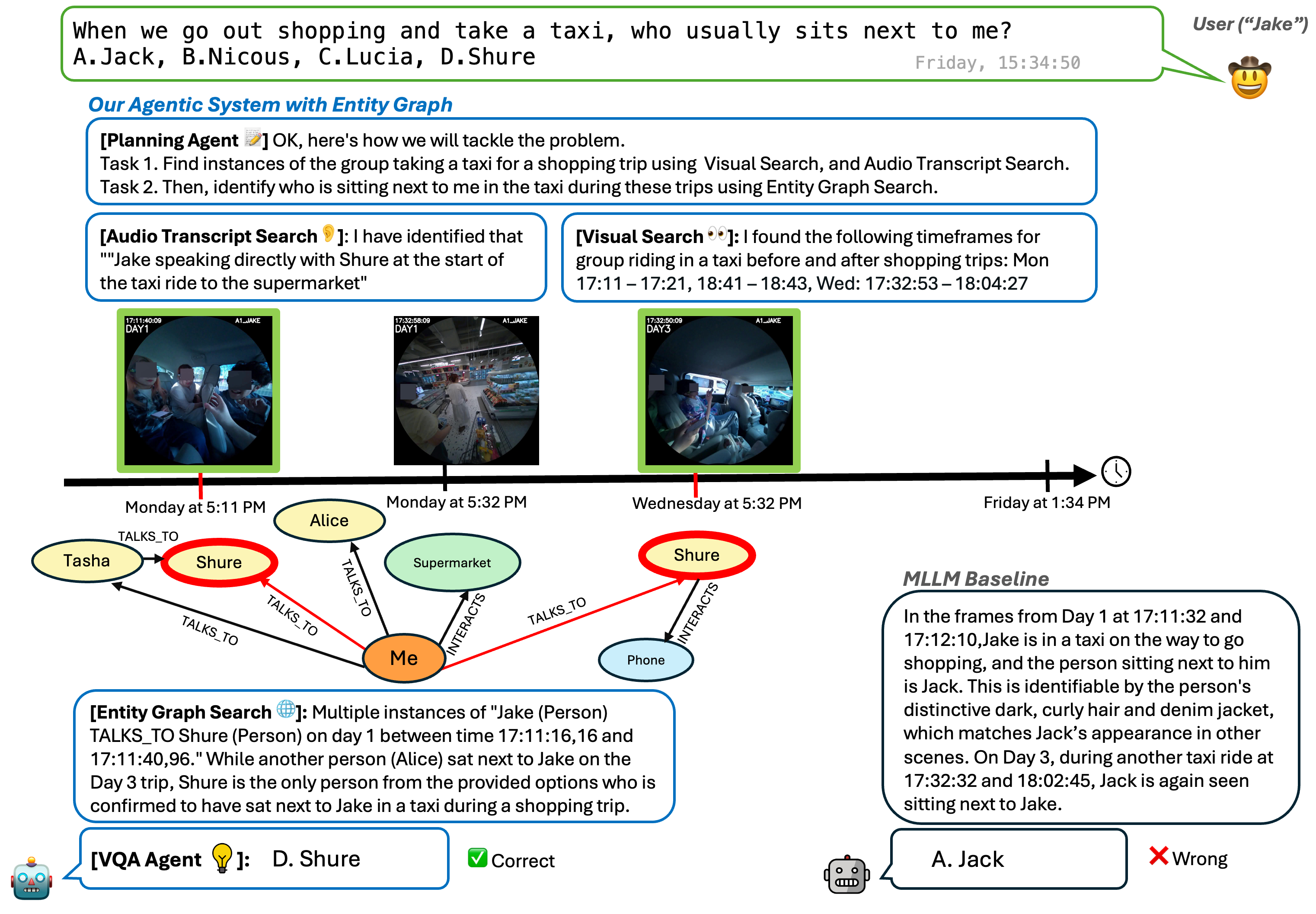}
\vspace{-4mm}
\caption{Given a natural language query, our agentic framework \mdlname{} decomposes the query into subtasks and leverages visual search, audio transcript search, and entity scene graph search to identify relevant events spanning multiple days. This example highlights the framework's ability to perform multi-hop, cross-modal reasoning by first performing temporal localization using audio and visual cues, and then using the entity graph to infer the answer. The entity graph consists of nodes for \emph{person}, \emph{object}, or \emph{location}, and edges capturing relations such as \emph{talks-to} and \emph{interacts-with}, each annotated with temporal intervals on when the relation holds.
}
\label{fig:teaser}
\end{center}

\section{Introduction}
\label{sec:intro}
Unlocking always-on personal AI assistants requires understanding not just isolated events, but a continuous stream of evolving user experiences. The recent emergence of AI-equipped wearable consumer devices such as the Ray-Ban Meta glasses, Amazon Echo Frames and Snapchat Spectacles as well as various prototypes~\citep{engel2023project, xu2025designing} creates an opportunity for AI agents to maintain persistent access to what users see and do over time.
For such assistants to provide helpful, personalized, and context-aware assistance, they need to possess \emph{longitudinal video understanding}, \emph{i.e.}, the ability to recall and interpret a user's lived experience over extremely long periods of time (days and months).

In this work, we address the challenge of ``very long video understanding''.
In prior literature, the definition of ``long'' has been continuously evolving. 
Popular benchmarks like MSR-VTT \citep{xu2016msrvtt} and DiDeMo \citep{hendricks2017didemo} where videos are up to a minute in length were once considered long, but recent works have further pushed this frontier to several minutes \citep{Wu2024LongVideoBenchAB, grauman2022ego4d} and up to an hour \citep{fu2025videomme, Zhou2024MLVUBM, wang2025lvbench}. The recent EgoLife \citep{yang2025egolife} pushes this frontier to \textbf{beyond 50 hours} of egocentric video over the course of a week, which is the length we define as \textbf{very long}. Unlike previous benchmarks that focus on large numbers of short, independent videos, EgoLife offers continuous, \textit{longitudinal} first-person video from six individuals.  This week-long horizon enables new research directions, such as tracking entities and their interactions across multiple days, analyzing repeated behaviors and habits, and handling extended periods of inactivity or ``lulls'' in the video stream. Agentic approaches, which equip agents with tools to search, retrieve, and reason over large corpora, have shown potential in addressing some of these limitations \citep{fan2024videoagent, wang2024videoagent, ma2025drvideo, chu2025graphvideoagent}. Existing agentic approaches often struggle to maintain coherent reasoning about entities and their relationships over extended temporal horizons, and have difficulty with fine-grained temporal localization such as tracking repeated actions or habits across days (\emph{e.g.}, ``how often did I drink water this week?''). Importantly, effective linkage between information from different modalities is needed to support richer and more accurate reasoning.

To address these challenges, we propose \textbf{\mdlname{}}, an enhanced agentic approach that centers on the extraction and use of an \textbf{entity scene graph} from long videos, where nodes represent people, places, and objects, and edges capture their relationships (\emph{e.g.}, uses, interacts with, mentions, talks to). Each edge is annotated with temporal intervals indicating when the relation holds.
In our proposed \mdlname{} system, we equip a planning agent with the ability to search and reason over this entity graph, as well as utilize a visual search tool (SQL + semantic search hybrid) and an audio transcript search tool. As illustrated in \cref{fig:teaser}, the system uses this graph in combination with audio and visual search to locate all shopping-related taxi rides across multiple days and infer who consistently sits next to the user.
By leveraging structured representations like entity graphs, \mdlname{} preserves complex relationships and supports detailed, compositional reasoning over extended timeframes, overcoming the limitations of existing methods. 

We evaluate our \mdlname{} pipeline on the EgoLifeQA benchmark and demonstrate state-of-the-art performance. Notably, \mdlname{} surpasses the previous state-of-the-art by 32\% and 39.7\% on the RelationMap and TaskMaster categories respectively, both of which require multi-hop relational reasoning. \mdlname{} also achieves competitive results on the Video-MME (Long) benchmark.

To summarize, our contributions are as follows:
\begin{itemize}
    \item We introduce an entity graph representation (\cref{ssec:eg_rep}) for long video understanding (\cref{ssec:task_setup}), enabling structured, cross-modal reasoning over very long time horizons.
    \item We present an agentic framework (\cref{ssec:agentic_framework}) that queries the entity graph along with visual and audio search tools, exceeding previous state-of-the-art performance on EgoLifeQA by 20.6\% (\cref{ssec:experiments_egolife}).
    \item We perform a detailed ablation study on entity graph construction and agentic tool usage for very long video understanding on EgoLife (\cref{ssec:ablations} and \Cref{app:egolife})
\end{itemize}
\section{Related Work}
\label{sec:relwork}

\noindent \textbf{Long Video Understanding with LLMs.}
The primary challenge in long-video understanding arises from the limited context window of large language models (LLMs), which restricts the amount of visual information processed at once. To address this, prior work focuses on condensing video inputs before LLM inference \citep{tang2025adaptive, lu2025decafnet, liu2025bolt}. Frame selection methods reduce input length by retaining only salient frames while preserving key content \citep{wang2025videotree, buch2025flexible, ye2025tstar}, whereas visual token compression techniques distill videos into compact token representations that better fit within context limits \citep{shen2025longvu, shu2025video}. These approaches can be query-dependent, selecting frames or tokens based on the input query \citep{liu2025hybrid, hu2025m, man2025adacm, diko2025rewind}, or query-independent, producing general summaries irrespective of downstream tasks \citep{yang2025pvc, zhao2025accelerating}. Other methods adopt sliding-window or hierarchical summarization strategies to maintain long-range context under fixed token budgets \citep{lu2025vited, zhou2024streaming}, or to directly extend the context capacity of LLMs themselves \citep{ding2024longrope, liu2023tcra, jin2024llm}.

\noindent \textbf{Video Understanding with Graph-based RAG.}
Retrieval-augmented generation (RAG) mitigates the context limitations of LLMs by retrieving relevant information from external sources \citep{lewis2020rag, gao2023ragsurvey}, which has also been extended to multimodal documents and long-video understanding \citep{yuvis2025rag, faysse2025colpali}. Traditional RAG operates over isolated text chunks, often losing relational context. To address this, Graph-based RAG methods such as GraphRAG \citep{edge2024local} and LightRAG \citep{guo2024lightrag} leverage knowledge graphs built from extracted entities and relations from the text corpus. More recently, researchers have begun to explore multi-modal RAG approaches, such as retrieving image frames directly instead of retrieving pre-generated video captions \citep{reddy2025video, wan2025clamr}. This approach preserves visual details that may be lost in textual abstraction, enabling more precise and comprehensive responses to complex queries. For instance, Video-RAG \citep{luo2025videorag} performs multi-modal RAG on video frames, automatic speech recognition (ASR) results, optical character recognition (OCR) results, and object-detection results. However, directly retrieving frames also introduces new challenges, including the need for efficient and accurate indexing, retrieval mechanisms, and effective data representations \citep{reddy2025video, wan2025clamr}. VideoRAG \citep{ren2025videorag} combines text-, visual-, and graph-based clip retrieval, matching queries to entity descriptions within a graph. AdaVideoRAG \citep{xue2025adavideorag} adaptively selects between no retrieval, naive retrieval, and graph-based retrieval based on question difficulty. RAVU \citep{malik2026ravu} uses VLMs to detect entities, generate frame descriptions, build spatio-temporal graphs, and infer answers. GraphVideoAgent \citep{chu2025graphvideoagent} iteratively retrieves relevant frames via caption-derived graphs. VideoMindPalace \citep{huang2025building} constructs layered spatio-temporal graphs encoding indoor layouts and activity zones, though its reliance on room-level structure limits robustness in open-ended scenes.

Many of these methods either overlook temporal relationships or construct graphs for the entire video at once. In contrast, we introduce an entity graph where each node is annotated with temporal information, making the graph time-aware and allowing it to be incrementally constructed as new data arrives. Experimentally, our method matches the performance of AdaVideoRAG \citep{xue2025adavideorag} on Video-MME (Long) while processing over ten times fewer frames. 

\noindent \textbf{Agentic Video Understanding.}
Recent advances in agentic video understanding have focused on developing systems that can autonomously perceive, reason, and act based on video content \citep{chen2025lvagent}. VideoAgent \citep{wang2024videoagent} introduces an agent-based framework where the agent is tasked with iteratively finding the relevant frames in the video for VQA if the information in the initial frames is not sufficient to answer the question. 
VideoAgent \citep{fan2024videoagent} iteratively employs tools such as object memory search and video-segment search based on video captions and visual embeddings to reach an answer.
DrVideo \citep{ma2025drvideo} reframes long-video understanding as long-document understanding by converting videos into text documents, iteratively augmenting them with key frame information and agent-based searches until enough information is gathered for chain-of-thought prediction. Similarly SiLVR \citep{zhang2026silvr} operates in the text domain by compressing dense visual captions and using a downstream reasoning LLM for video understanding.%

Our proposed \mdlname{} advances agentic video understanding by integrating a temporally-annotated entity scene graph into the tool-calling loop. Unlike prior systems that rely on unstructured captions or repeated frame retrieval, our approach enables efficient cross-modal search and compositional reasoning for complex, longitudinal queries.
\section{Method}
\label{sec:method}
Here we formalize the task of very long video understanding (\cref{ssec:task_setup}) and describe extracting entity graph representations of such long videos (\cref{ssec:eg_rep}). Lastly, we discuss the design of the proposed agentic framework \mdlname{} which utilizes these entity graph representations for very long video understanding (\cref{ssec:agentic_framework}).

\subsection{Task Setup}
\label{ssec:task_setup}

We focus on the task of very long video understanding, specifically on video question-answering over videos that potentially span an entire week.
Let $\mathcal{V} = \{v_t\}_{t=1}^T$ denote the video sampled at 1 FPS (frame per second). 
Similarly, let $\mathcal{AT} = \{u_i, t_{start_i}, t_{end_i}\}_{i=1}^N$ denote the set of transcribed speech $u_i$ with associated time-stamps $(t_{start_i}, t_{end_i})$. At test time, the system receives a complex query $Q$ in natural language, and must produce a textual answer $A$. 
Formally, the task is to obtain a mapping $H: (\mathcal{V}, \mathcal{AT}, Q) \xrightarrow{} A$.

Naively feeding all frames and transcripts into a multimodal LLM or VLM for such very long videos is infeasible due to context window limitations.
The prevailing approach, Video Retrieval Augmented Generation (RAG) \citep{luo2025videorag}, first selectively retrieves a small subset of frames and audio transcripts deemed relevant to the user query $Q$ and conditions the VLM on this retrieved set to generate the answer $A$. 
However, a naive RAG approach over very long egocentric videos is insufficient to answer egocentric queries which are often entity-centric and require multi-hop reasoning across days. 
These include tracking repeated behaviors, or interactions between specific people, objects, and locations. 
Direct embedding-based retrieval over unstructured clips or captions struggles to maintain coherent entity identities over time to support compositional constraints such as ``all times I talked to person X this week''.

We address this in two steps. 
First, to support queries over entity relations over time, we construct an entity-centric scene graph that explicitly encodes people, objects, locations, temporally localized relations, and provide a structured index to allow narrowing down to the relevant regions of the video (\cref{ssec:eg_rep}).
Second, we propose an agentic framework \mdlname{} which involves a planning agent that iteratively decomposes $Q$ into sub-tasks and invokes specialized retrieval tools including the above constructed entity graph (\cref{ssec:agentic_framework}). 

\begin{figure*}[t]
\centering
\includegraphics[width=\linewidth]{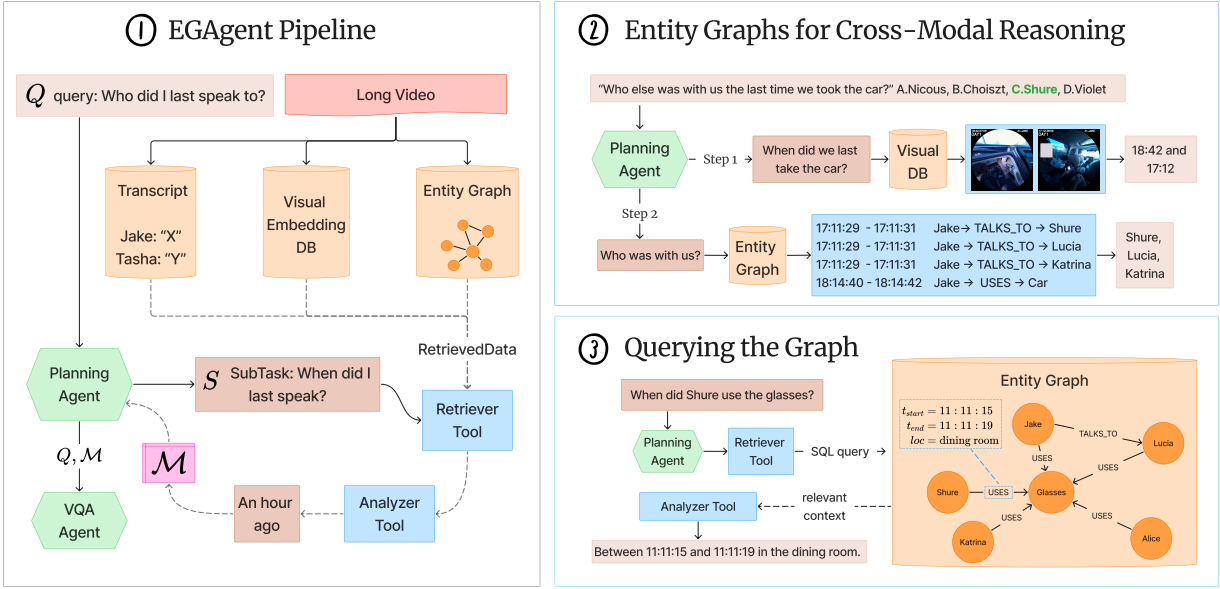}
\vspace{-6mm}
\caption{We show an overview of our \mdlname{} pipeline for very long video understanding using cross-modal reasoning in \textcircled{1}. Given a very long video and a query, a planning agent devises a multi-step plan of sub-tasks required to answer the query. The planning agent uses a retriever tool to probe three data sources extracted from the long video: audio transcripts, visual frame embeddings, and an entity scene graph, which is the focus of \mdlname{}. We show an example of how the planning agent composes cross-modal information retrieved from the visual database and entity graph to answer an EgoLife query in \textcircled{2}. We visualize the entity graph query mechanism in \textcircled{3}, where the retriever tool designs a SQL query to retrieve relevant relationships for the planning agent to reason over.}
\vspace{-4mm}
\label{fig:pipeline}
\end{figure*}

\subsection{Entity Graph Representations}
\label{ssec:eg_rep}
From our observations, baseline methods often struggle with questions that require understanding a person's habits or repeated behaviors over time (\emph{e.g.}, ``What do I often check on my phone in the morning?''), as well as those that involve reasoning about interactions and relationships between different entities, such as people, objects, or places across extended periods (\emph{e.g.}, ``Before we went to see the dog, who went with me to the second floor to find Tasha?''). 
Because these methods do not explicitly model entity relationships or track long-term behavioral patterns, their performance on such questions, especially over long time horizons, is limited. 

To address this, we construct an entity graph $G=(V,E)$ to capture relationships and interactions, enabling the planning agent to query this graph during inference.

\begin{itemize}
    \item \textit{Nodes ($V$)}: entities (\emph{i.e.}, individuals, objects, places)
    \item \textit{Edges ($E$)}: relationships (\emph{i.e.}, interacts with, mentions, talks to, uses), and temporal information
\end{itemize}

\begin{figure}
\centering
\includegraphics[width=0.9\linewidth]{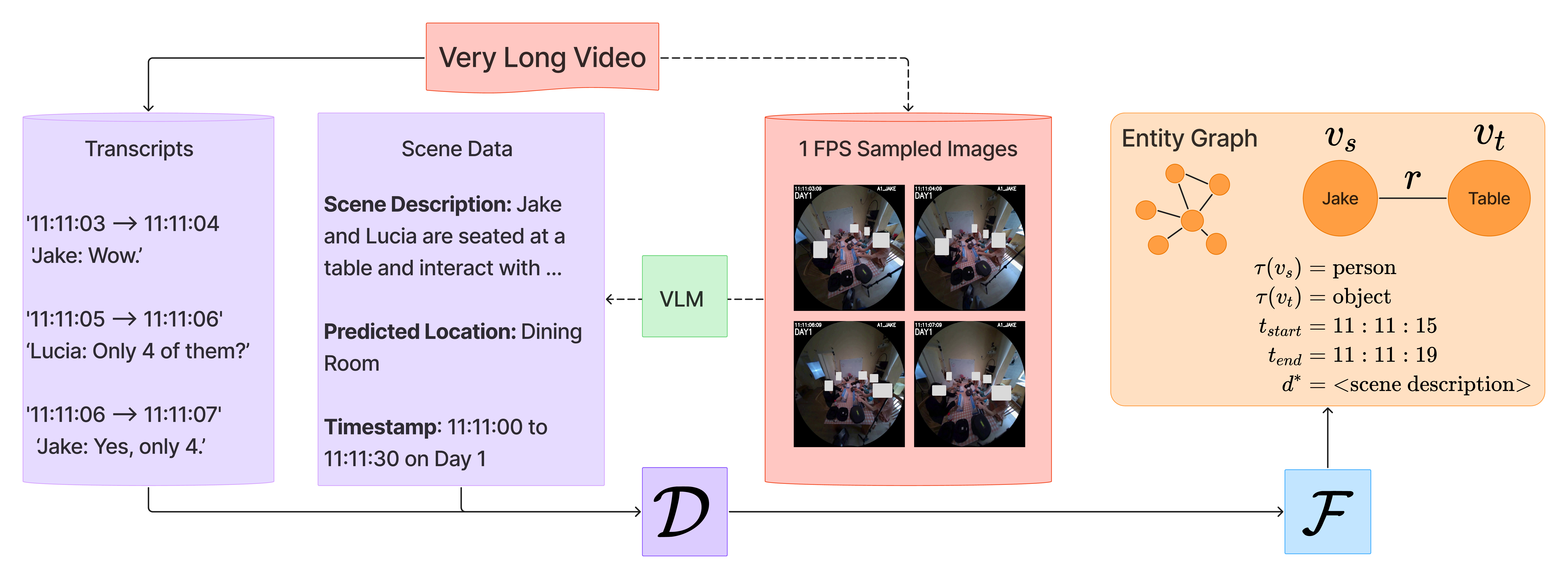}
\vspace{-2mm}
\caption{We use an LLM, denoted as $\mathcal{F}$, to extract an entity graph from text documents $\mathcal{D}$ that represent a very long video, \emph{i.e.}, audio transcripts $\mathcal{AT}$ and scene descriptions and locations extracted from sampled image frames $\mathcal{V}$ (see \Cref{app:implementation} for details). Each graph relationship $r$ connects a source vertex $v_s$ and target vertex $v_t$ between time $(t_\mathrm{start}, t_\mathrm{end})$. Each vertex has an entity type $\tau(v)$ and the raw text document $d^*$ used to extract the relationship (\cref{ssec:agentic_framework}).}
\label{fig:eg_creation}
\vspace{-2mm}
\end{figure}

Each edge is annotated with temporal information, allowing us to track the existence, sequence and duration of the corresponding relationships. 
Such temporal structure is crucial for reasoning about events and interactions that unfold or repeat across long horizons.

\noindent 
\textbf{Entity Graph Creation.}
We construct an entity graph $G {=} (V, E)$ from a given collection of text documents $\mathcal{D}$ which includes audio transcripts, scene descriptions, predicted scene locations (illustrated in \cref{fig:eg_creation}). We discuss details of extracting scene data to generate these documents $\mathcal{D}$ in \Cref{app:implementation}. For each document $d \in \mathcal{D}$, we apply an LLM-based extractor $\mathcal{F}$ to jointly identify entities and their relationships:
\begin{equation}
(V_d, E_d) = \mathcal{F}(d)
\label{eq:node_edge_extraction}
\end{equation}
Here, $V_d$ is the set of entities and $E_d$ is the set of relationships extracted from $d$.  
The overall entities and relationships are aggregated as:
 \begin{equation}
(V, E) = \left(~ \bigcup_{d \in \mathcal{D}} V_d,\; \bigcup_{d \in \mathcal{D}} E_d \right)
\label{eq:aggregate_entities_edges}
\end{equation}
    We assign each node $v {\in} V$ a type $\tau(v)$ to be one of ``person'', ``object'', ``location''.    
    We initially represent each edge $e$ as a tuple $(v_s, v_t, r)$, where $v_s$ and $v_t$ are the source and target nodes, and $r \in \mathcal{R}$ is the relationship type. The set of relationship types is:
\begin{equation}
\mathcal{R} = \{\mathrm{talks\!-\!to},\mathrm{interacts\!-\!with},\mathrm{mentions},\mathrm{uses}\}
\label{eq:relationship_types}
\end{equation}

    Each edge $e$ is subsequently annotated with temporal information $(t_\mathrm{start}, t_\mathrm{end})$ derived from the source document $d$. 
    After temporal annotation, each edge is represented as: 
 \begin{equation}
e = (v_s, v_t, r, t_\mathrm{start}, t_\mathrm{end})
\label{eq:edge_representation}
\end{equation}
    
The resulting graph is stored as a set of tuples:
\begin{equation}
(v_s, \tau(v_s), v_t, \tau(v_t), r, t_\mathrm{start}, t_\mathrm{end}, d^*)
\label{eq:graph_storage_tuple}
\end{equation}
$d^*$ is the supporting text snippet from which the edge was extracted. 
The graph is stored in memory as a SQLite3 database, with each row corresponding to one tuple. 
The graph construction process supports incremental updates as new documents $d$ arrive, allowing $G$ to grow and refine over time.

\begin{algorithm}[t]
\caption{\mdlname{} Framework}
\label{alg:agentic_framework}
\begin{algorithmic}[1]
\REQUIRE User query $Q$, Multimodal data sources (Video, Audio, Entity Graph)
\ENSURE Final answer $A$
\STATE Initialize working memory $\mathcal{M} \gets \emptyset$
\STATE // \textbf{Step 1: Joint Decomposition and Tool Selection}
\STATE SubtaskList $\gets$ \texttt{PlanningAgent.decompose\_and\_select}($Q$) \\
\COMMENT{SubtaskList = $\{(S_1, T_1,  q_1), (S_2, T_2, q_2), \ldots, (S_N, T_N, q_N)\}$}
\FOR{each $(S, T, q)$ in SubtaskList}
    \STATE // \textbf{Step 2a: Retrieve relevant data for the subtask}
    \STATE RetrievedData $\gets$ $T(q)$ 
    \COMMENT{Visual: hybrid semantic/attribute search; Audio: transcript search; Entity Graph: SQL queries}
    \STATE // \textbf{Step 2b: Analyze retrieved data for relevance and evidence}
    \STATE Analysis $\gets$ \texttt{AnalyzerTool.analyze}(RetrievedData, $S$)
    \COMMENT{LLM-based reasoning, evidence extraction, filtering}
    \STATE // \textbf{Step 2c: Update working memory}
    \STATE $\mathcal{M} \gets \mathcal{M} \cup \{\text{Analysis}\}$
\ENDFOR
\STATE // \textbf{Step 3: Final Synthesis}
\STATE $A \gets \texttt{VQAAgent.answer}(Q, \mathcal{M})$
\COMMENT{VQAAgent uses accumulated, cross-modal evidence in $\mathcal{M}$ to answer $Q$}
\RETURN $A$
\end{algorithmic}
\end{algorithm}

\subsection{Agentic Framework \mdlname{}}
\label{ssec:agentic_framework}
Given the very long video and entity graph representation described above, we propose an agentic framework \mdlname{} for multi-modal reasoning, summarized in \cref{alg:agentic_framework} and illustrated in \cref{fig:pipeline}.
\mdlname{} consists of six main components: a \textbf{Planning Agent}, three \textbf{Retriever Tools} (Visual Search, Audio Transcript Search, and Entity Graph Search), an \textbf{Analyzer Tool}, and a \textbf{VQA Agent} (see \textcircled{1} in~\cref{fig:pipeline}). We discuss more details of our agent design and provide qualitative examples in~\cref{app:agent_design}.

Each component operates over a specific data modality or reasoning step. 
The Planning Agent decomposes a complex user query $Q$ into sub-tasks, selects appropriate tools, and maintains a working memory $\mathcal{M}$ that accumulates cross-modal evidence. 
Retriever Tools (Visual Search,  Audio Transcript Search, Entity Graph Search) access different data sources to find relevant information for each sub-task, the Analyzer Tool filters and distills retrieved information, and the VQA Agent produces the final answer $A$ from the accumulated evidence.

\textbf{Planning Agent} orchestrates the entire reasoning process. Given a user query $Q$ along with natural language definitions for each tool, the Planning Agent performs a joint decomposition of $Q$ into a sequence of $N$ sub-tasks $\{S_1, S_2, \ldots, S_N\}$, each sub-task with an associated retriever tool $T_i$ and query arguments $q_i$ (Lines 2--3 in~\Cref{alg:agentic_framework}).
When a sub-task requires multiple modalities, the planner may emit consecutive
planner steps that repeat the same sub-task description, each routed to a
different tool, e.g., $(S_i, T_\mathrm{vis}, q_i)$, $(S_i, T_\mathrm{eg}, q_{i+1})$,
$(S_i, T_\mathrm{aud}, q_{i+2})$.

Each sub-task $S_i$ targets a specific aspect of the information needed such as object localization, checking diarized speech, or confirming past interaction. 
Each tuple $(S_i, T_i, q_i)$ invokes one of the following retriever tools:
(i) \textbf{Visual Search Tool} ($Tool_\mathrm{vis}$) retrieves visual content.  
(ii) \textbf{Audio Transcript Search Tool} ($Tool_\mathrm{aud}$) retrieves transcribed speech. 
(iii) \textbf{Entity Graph Search Tool} ($Tool_\mathrm{eg}$) queries an entity-centric scene graph.
The retrieved content is passed to the \textbf{Analyzer Tool} and the corresponding analysis is updated to the working memory $\mathcal{M}$. 
Such an iterative process allows \mdlname{} to progressively refine its understanding of the query $Q$ while keeping per-sub-task context size manageable.
Finally, the \textbf{VQA Agent} consumes the working memory and original query to provide a final answer. See \textcircled{2} in~\cref{fig:pipeline} for an example of how the planning agent reasons over cross-modal information retrieved with retriever tools.

\textbf{Visual Search Tool} samples video frames at 1FPS and embeds 
each frame $v_t$ as $\phi_I(v_t) \in \mathbb{R}^d$ using a vision-encoder \citep{tschannen2025siglip}. 
The generated embeddings along with attributes such as timestamp, location are stored in a SQLite database which supports efficient retrieval. 
At inference, the Planning Agent provides a text sub-query $q_i$ (embedded as $\phi_T(q)$)  and optional attribute filters $f$ (\emph{e.g.}, ``kitchen", ``morning'').
The tool computes cosine similarity $\cos(\phi_T(q), \phi_I(v_t))$ for filtered rows in the SQLite database returning the $k$-nearest neighbors for further analysis.

\textbf{Audio Transcript Search Tool} operates over text transcripts. 
 We consider two variants (i) LLM-based search where we feed entire transcripts to an LLM for a relevant time range (parallelized over days due to context limits) (ii) BM25-based lexical search. 
 The former provides significantly better quality results at the cost of higher latency. 

\textbf{Entity Graph Search Tool} queries the entity-centric scene graph $G$ introduced in \cref{ssec:eg_rep} and stored tuples in a SQLite database (\Cref{eq:graph_storage_tuple}).
During inference, the Planning Agent issues SQL queries $q$ over the following fields: (i) time filter (ii) keyword text search (iii) entity  source and/or target nodes $(v_s, v_t)$ and (iv) relationship type $r$. 
In practice, real-world data is often incomplete or noisy, so the Planning Agent adopts a ``strict-to-relaxed'' query strategy: it first issues an exact match query on all specified fields, and if no results are found, incrementally relaxes constraints by broadening the time window, allowing partial text matches, and finally relaxing the relationship type filter. 
This strategy maximizes precision when possible while increasing recall when exact matches are unavailable (see \textcircled{3} in~\cref{fig:pipeline} for an example query trace and \cref{app:eg_extraction} for qualitative examples of SQL querying).

\textbf{Analyzer Tool} determines the relevance of the retrieved context for each sub-task $S_i$ via an LLM to perform lightweight reasoning, evidence extraction, and optional de-duplication. 

\textbf{VQA Agent} is a multi-modal LLM that conditions on $Q$ and the compact evidence in $\mathcal{M}$ to generate the final answer $A$ (Algorithm~\ref{alg:agentic_framework}, Line 13), enabling detailed, temporally coherent reasoning over week-long egocentric videos.
\section{Experiments}
\label{sec:experiments}
We evaluate \mdlname{} against baselines on two benchmarks, EgoLifeQA and Video-MME (Long), which focus on \textit{very long} video understanding. Here we discuss implementation details of \mdlname{} and analyze its performance on these datasets. Lastly, we discuss a few ablation studies on entity graph extraction and wall-clock latency.

\subsection{Evaluation Benchmarks}
 \noindent 
\textbf{EgoLifeQA:} EgoLifeQA consists of 500 long-context Multiple-Choice Questions (MCQs) derived from the EgoLife~\citep{yang2025egolife} dataset, in which six participants lived together for one week, continuously recording their daily activities using Project Aria glasses~\citep{engel2023project}. 
The benchmark focuses on the 50 hours of videos taken from the perspective of Jake, one of the six participants. 
The MCQs cover practical questions such as locating items, recalling past events, tracking habits, and analyzing social interactions. 
Each question has four candidate answers with a single correct option. 
Each question is associated with \textit{query time} (\emph{e.g.}, 11:34 AM on day 4) and a manually verified \textit{target time}, indicating the specific portion of the video that contains the information needed to answer the MCQ correctly.

\noindent 
\textbf{Video-MME (Long):} Video-MME~\citep{fu2025videomme} comprises 900 videos, with 2700 MCQs. The benchmark is divided into \textit{Short}, \textit{Medium}, and \textit{Long} subsets based on video length. 
We focus on the \textit{Long} subset that consists of 300 videos that range from 30 to 60 minutes. (\cref{ssec:experiments_videomme}).

\subsection{Implementation Details}
To prepare the entity graphs for our experiments, we extract a separate graph for each video in the Video-MME~\citep{fu2025videomme} dataset. 
For EgoLifeQA~\citep{yang2025egolife}, due to the increased likelihood of LLM invocation failures with longer input transcripts, we instead extract one graph per hour of video.
In both datasets, audio is represented by text transcripts. For Video-MME, transcripts are generated using an ASR foundation model such as Whisper. In contrast, EgoLife provides manually diarized transcripts, which include both speaker identities and the corresponding speech content. We discuss more details and provide code snippets and all agent and tool use prompts in \Cref{app:implementation}.

\begin{table*}[t!]
\centering
\caption{MCQ Accuracy on EgoLifeQA \citep{yang2025egolife}. The previously reported state-of-the-art is underlined, the current state-of-the-art is bolded, and the current second-best italicized. Agentic approaches are given frames or captions sampled at 1FPS and then choose a subset X for analysis, which is denoted by 1FPS$\to$X under \# Frames. F = raw video frames, C = video captions, A = raw audio, T = audio transcript. ``--'' in results of individual categories denotes missing data as they were not reported in the original papers. We estimate token usage for these baselines, which are marked with an asterisk* (see \Cref{app:implementation} for details on estimation). The following are question type categories from EgoLifeQA, for which we report MCQ Accuracy below: EL (EntityLog), ER (EventRecall), HI (HabitInsight), RM (RelationMap), TM (TaskMaster).}
\vspace{-2mm}
\resizebox{\textwidth}{!}{%
\begin{tabular}{@{}c|l|c|c|cccccc|c|c@{}}
\toprule
\multirow{2}{*}{\textbf{Category}}                                                  & \multicolumn{1}{c|}{\multirow{2}{*}{\textbf{Method}}}                   & \multirow{2}{*}{\textbf{\# Frames}} & \multirow{2}{*}{\textbf{Modality}} & \multicolumn{6}{c|}{\textbf{MCQ Accuracy (\%)}}                                                                                         & \multirow{2}{*}{\textbf{\begin{tabular}[c]{@{}c@{}}Average\\ Gain (\%)\end{tabular}}} & \multirow{2}{*}{\textbf{\begin{tabular}[c]{@{}c@{}}Average\\ \# Tokens\end{tabular}}} \\ \cmidrule(lr){5-10}
& \multicolumn{1}{c|}{}                                                   &                                     &                                    & \textbf{EL} & \textbf{ER} & \textbf{HI} & \textbf{RM} & \textbf{TM} & \textbf{Average} &                                                                                       &                                                                                       \\ \midrule
\multirow{3}{*}{\begin{tabular}[c]{@{}c@{}}MLLMs\\ (Uniform \\  Sampling)\end{tabular}} & LLaVA-Video-7B        & 64                                  & F                                  & --                 & --                   & --                    & --                   & --                  & 36.4             &                                                                                       & ~~~~32K*                                                                                    \\
& GPT-4.1                                                               & 1FPS                                & C                                  & 32.0               & 39.7                 & 39.3                  & 32.8                 & 39.7                & 36.0             &                                                                                       & 285K                                                                                  \\
& Gemini 2.5 Pro                                                   & 3000                                & F, T                               & 45.6               & 48.4                 & 51.7                  & 41.6                 & 52.4                & 46.8             & +9.9                                                                                  & 807K                                                                                  \\ \midrule
RAG                                                                                 & LLaVA-Video-7B + Video-RAG                     & 64                                  & F                                  & --                 & --                   & --                    & --                   & --                  & 30.0             &                                                                                       & ~~~~18K*                                                                                    \\ \midrule
\multirow{5}{*}{\begin{tabular}[c]{@{}c@{}}Agentic\\ Baselines\end{tabular}}        & EgoButler Gemini 1.5 Pro                        & 0                                   & C, T                               & 36.0               & 37.3                 & 45.9                  & 30.4                 & 34.9                & {\ul 36.9}       & +0                                                                                    & ~~~~26K*                                                                                     \\
& EgoButler GPT-4o                                & 0                                   & C, T                               & 34.4               & 42.1                 & 29.5                  & 30.4                 & 44.4                & 36.2             &                                                                                       & ~~~~19K*                                                                                    \\
& VideoAgent                               & 1FPS$\to$8                          & F                                  & --                 & --                   & --                    & --                   & --                  & 29.2             &                                                                                       & ~~128K*                                                                                    \\
& LLaVA-OneVision-7B + T* & 1FPS$\to$8                          & F, T                               & --                 & --                   & --                    & --                   & --                  & 35.4             &                                                                                       & ~~~~32K*                                                                                    \\
& Ego-R1 Qwen-2.5-3B-Instruct                       & 1FPS                                & F, C, T                            & --                 & --                   & --                    & --                   & --                  & 36.0        &                                                                                       & ~~128K*                                                                                    \\ \midrule
\multirow{4}{*}{Ours}                                                               & \mdlname~ GPT-4.1 (F + T)                                                  & 1FPS$\to$50                         & F, T                               & {48.0}         & 48.4                 & 55.7                  & 40.0                 & 61.9                & 48.6             &     +11.7                                                                                  &  551K                                                                                     \\
& \mdlname~ GPT-4.1 (EG + F + T)                                             & 1FPS$\to$50                         & F, C, T                            & 44.0               & 49.2                 & 55.7                  & {53.6}           & {66.7}          & {50.7}       &  +13.8                                                                                     &  571K                                                                                     \\ 
& \mdlname~ GPT-4o (EG + F + T)                                              & 1FPS$\to$50                         & F, C, T                            & 44.8               & \textit{54.8 }                & \textit{59.0}                  & 44.0                 & 61.9                & 44.6             & ~~+7.7                                                                                  & 652K                                                                                  \\
& \mdlname~ Gemini 2.5 Pro (EG + F + T)                                           & 1FPS$\to$50                         & F, C, T                            & \textbf{54.4}      & \textbf{57.1}        & \textbf{60.3}         & \textbf{62.4}        & \textbf{74.6}       & \textbf{57.5}    & \textbf{+20.6}                                                                        & 880K                                                                                  \\ \bottomrule
\end{tabular}
}
\label{tab:egolifeqa_results}
\vspace{-2mm}
\end{table*}
\begin{figure}[th!]
\centering
\includegraphics[width=.7\linewidth]{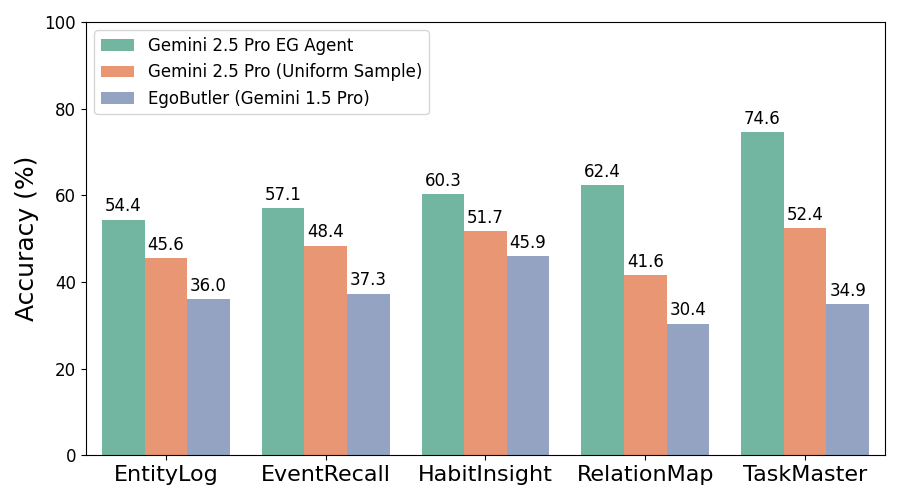}
\caption{The performance comparison against Gemini 2.5 Pro and EgoButler in each question category in EgoLifeQA. Our approach significantly outperforms baselines on RelationMap (+20.8\%) and TaskMaster (+22.2\%), where entity understanding and complex reasoning is required to provide a correct answer.}
\label{fig:egolife_category_results}
\vspace{-2mm}
\end{figure}

\subsection{Analysis on EgoLifeQA Benchmark}
\label{ssec:experiments_egolife}
We compare the proposed \mdlname{} against strong baselines in three categories: 1) MLLM with uniform sampling; 2) MLLM with RAG; and 3) existing agentic approaches.

\noindent 
\textbf{Baselines.} To handle extremely long videos in EgoLifeQA, frame sampling in MLLM baselines varies based on their respective context window size.
GPT-4.1 takes video captions that were generated for every 30-second video snippet sampled at 1 FPS.
We sample 3000 frames uniformly along with the audio transcripts for Gemini 2.5 Pro. 
The results of LLaVA-Video-7B~\citep{zhang2025llavavideo} and LLaVA-Video-7B combined with Video-RAG~\citep{luo2025videorag} are reported in~\citet{yang2025egolife}.

We compare our \mdlname{} with the following existing agentic methods: EgoButler~\citep{yang2025egolife}, a hierarchical text-based Retrieval-Augmented Generation (RAG) approach, and Ego-R1~\citep{tian2025egor1}, a lightweight 3B-parameter agent trained on egocentric data, including portions of EgoLife for tool calling. 
We report results of all RAG and existing agentic approaches in~\cref{tab:egolifeqa_results} directly from these works.

\noindent 
\textbf{Performance Analysis.}
\cref{tab:egolifeqa_results} presents a comprehensive comparison of methods on the EgoLifeQA benchmark. 
Our proposed agentic system, which incorporates entity graph reasoning, achieves strong performance across all evaluation categories and establishes a new state-of-the-art. 
Notably, while Gemini 2.5 Pro with uniform sampling already outperforms the previous best results (EgoButler), our \mdlname{} with a Gemini 2.5 Pro backbone delivers an additional improvement of 10.7\%, highlighting the significant value of entity graph reasoning. 

Furthermore, the benefits of entity graph reasoning are not limited to a single MLLM backbone.
Applying the same agentic framework to the GPT-4.1 backbone also yields notable gains over its uniform sampling counterpart. 
These results demonstrate that integrating entity graph reasoning within agentic systems consistently enhances performance on very long video understanding tasks. 

To compare with existing agentic systems, we run \mdlname{} on the same LLM backbone (GPT-4o) with other agentic systems. Our entity graph agent consistently surpasses other agentic approaches utilizing the same model, including EgoButler (+8.4\%), VideoAgent (+15.4\%), and Ego-R1 (+8.6\%).

\cref{tab:egolifeqa_results} demonstrates that, when using the same backbone, the proposed method incorporating an entity graph substantially improves performance on EgoLifeQA. It outperforms the baseline (without entity graph) in 4 out of 5 categories, with particularly notable gains in the RelationMap and TaskMaster categories. This improvement can be attributed to the entity graph's ability to enable cross-modal reasoning.

\cref{fig:egolife_category_results} further illustrates the performance gap between the proposed agentic approach and Gemini 2.5 Pro with uniform sampling. It is evident that the agentic system benefits the most on RelationMap and TaskMaster categories, which require multi-hop relational reasoning. Specifically, our approach surpasses the previous state-of-the-art and Gemini 2.5 Pro by 32\% and 20.8\%, respectively, on RelationMap QAs, and achieves impressive gains of 39.7\% and 22.2\% in the TaskMaster category. We discuss more examples and benchmark analyses in~\cref{app:egolife}.

\begin{table*}[t!]
\centering
\tiny
\caption{MCQ Accuracy on Video-MME (Long). The current state-of-the-art is bolded and the second-highest is underlined. F = raw video frames, C = video captions, A = raw audio, T = audio transcript, O = object detection bounding boxes. ``1 FPS~$\rightarrow$~50'' denotes retrieving 50 frames sampled at 1 FPS which are used for MLLM analysis. We estimate token usage wherever unreported, which are marked with an asterisk* (see \Cref{app:implementation} for details on estimation).
}
\vspace{-1mm}
\footnotesize
\begin{tabular}{@{}c|l|ccc|c|c@{}}
\toprule
\textbf{Category}                                                            & \multicolumn{1}{c|}{\textbf{Method}}                               & \textbf{Context} & \textbf{\# Frames} & \textbf{Modality} & \textbf{Accuracy (\%)} & \textbf{\# Tokens} \\ \midrule
\multirow{2}{*}{\begin{tabular}[c]{@{}c@{}}MLLMs\\(Uniform Sampling)\end{tabular}}  & Gemini 2.5 Pro                      & 1M                                                                & 256                & F, A              & \textbf{82.0}     & ~~100K*                                                         \\
& GPT-4.1                                                           & 1M                                                                & 384                & F                 & 72.0              & ~~~~60K*                                                                 \\
\midrule
\multirow{2}{*}{RAG}                                                         
& Video-RAG (Qwen2.5-VL-7B)                 & 32K                                                               & 32                 & F, O, T           & 43.3              & ~~~~10K*                                                                   \\
& AdaVideoRAG (Qwen2.5-VL-7B)             & 128K                                                               & 768                & F, C, T           & 47.7              & ~~128K*                                                                   \\ \midrule
\multirow{2}{*}{\begin{tabular}[c]{@{}c@{}}Agentic\\ Baselines\end{tabular}} & DrVideo (DeepSeek V2.5)                      & 128K                                                              & 0.2 FPS            & F, T              & 71.7              & ~~128K*                                                                   \\ 
& \begin{tabular}[c]{@{}l@{}}VideoDeepResearch \\ (DeepSeek-r1-0528 + Qwen2.5VL-7B)\end{tabular} & 32K              & 32                 & F, T              & 72.4              & ~~~~32K*                  \\ 
\midrule
\multirow{2}{*}{Ours}                                                        & \mdlname{} (Qwen2.5-VL-7B)                                          & 32K                                                               &  1FPS$\to$50                  & F, C, T           & 47.8                  & 172K                                                                  \\
& \mdlname{} (Gemini 2.5 Pro)                                         & 1M                                                                &  1FPS$\to$50                  & F, C, T           & {\ul 74.1}                   & 134K                                                                   \\ \bottomrule
\end{tabular}
\label{tab:videomme_results}
\end{table*}
\begin{table}[t!]
\centering
\caption{A comparison on EgoLifeQA of Entity Graph Extraction (EGX) using only transcript (T) vs a transcript-fused caption (C+T), and swapping out the transcript search tool from an LLM search to BM25 lexical search. All \mdlname{} methods reason over the entity graph, frames and audio transcripts (EG + F + T). EgoButler uses transcript-fused captions (C +T). All gains (\%) are with respect to EgoButler GPT-4o.}
\vspace{-1mm}
\resizebox{.6\columnwidth}{!}{%
\begin{tabular}{@{}l|cccccc@{}}
\toprule
\multicolumn{1}{c|}{\multirow{2}{*}{\textbf{Method}}} & \multirow{2}{*}{\textbf{VLM}}   & \multirow{2}{*}{\textbf{EGX}} & \multirow{2}{*}{\textbf{\# F}} & \multirow{2}{*}{\textbf{T Search}} & \multirow{2}{*}{\textbf{Accuracy (\%)}} & \multirow{2}{*}{\textbf{Gain (\%)}} \\
&                                 &                               &                                &                                    &                                         &                                     \\ \midrule
EgoButler                                            & GPT-4o                          & --                            & 0                              & LLM                                & 36.2                                    & -                                   \\ \midrule
\multirow{11}{*}{\mdlname{}}                         & \multirow{3}{*}{GPT-4o}         & T                             & 50                             & BM25                               & 36.6                                    & +0.4                                \\
&                                 & T+C                           & 50                             & BM25                               & 39.4                                    & +3.2                                \\
&                                 & T+C                           & 50                             & LLM                                & 44.6                                    & +8.4                                \\ \cmidrule(l){2-7} 
& \multirow{5}{*}{GPT-4.1}        & T                             & 0                              & -                                  & 36.8                                    & +0.6                                \\
&                                 & T                             & 50                             & BM25                               & 42.2                                    & +6.0                                \\
&                                 & T                             & 50                             & LLM                                & 49.2                                    & +13.0                               \\
&                                 & T+C                           & 50                             & BM25                               & 43.9                                    & +7.7                                \\
&                                 & T+C                           & 50                             & LLM                                & 50.7                                    & +14.5                               \\ \cmidrule(l){2-7} 
& \multirow{3}{*}{Gemini 2.5 Pro} & T                             & 50                             & BM25                               & 48.6                                    & +12.4                               \\
&                                 & T+C                           & 50                             & BM25                               & 51.8                                    & +15.6                               \\
&                                 & T+C                           & 50                             & LLM                                & 57.5                                    & +21.3                               \\ \bottomrule
\end{tabular}
}
\vspace{-2mm}
\label{tab:egolife_bm25_vs_llm_search}
\end{table}

\subsection{Analysis on Video-MME Benchmark}
\label{ssec:experiments_videomme}

We also evaluate our entity graph agent on the \textit{long} subset of Video-MME in \cref{tab:videomme_results}. 
Because Gemini 2.5 Pro can process native video (frames + audio) without the need for transcripts, it remains the state-of-the-art in this sub-hour length regime. Using an identical LLM backbone (Qwen2.5-VL-7B), \mdlname~ surpasses Video-RAG (+4.5\%), and matches the performance of AdaVideoRAG while processing over $10\times$ fewer frames.

Compared with recent agentic approaches~\citep{guo2025deepseek, yuan2025videodeepresearch} that use frontier models as their LLM backbone, our \mdlname{} with a Gemini 2.5 Pro backbone demonstrates strong performance, second only to native Gemini 2.5 Pro that processes 256 frames. In contrast, \mdlname{} uses only a fifth of the image frames compared to the baseline. 
More importantly, uniformly sampling with MLLMs like Gemini 2.5 Pro does not scale well to extremely long videos, as demonstrated in the EgoLifeQA benchmark~\cref{ssec:experiments_egolife}.

\begin{table}[t!]
\centering
\caption{Wall-clock runtime of \mdlname{} that reasons over the entity graph, frames and audio transcripts (EG + F + T) on EgoLifeQA.}
\resizebox{.5\columnwidth}{!}{%
\begin{tabular}{@{}c|cccc@{}}
\toprule
Method                              & T Search & Accuracy (\%) & Runtime (sec) & \#Tokens \\ \midrule
\multirow{2}{*}{\mdlname{} GPT-4.1} & BM25     & 43.9          & 125           & 172K     \\
& LLM      & 50.7          & 169           & 571K     \\ \bottomrule
\end{tabular}
}
\label{tab:quality_vs_latency}
\end{table}

\subsection{Ablations}
\label{ssec:ablations}

\noindent
\textbf{Extraction of Entity Graph.} We compare two variants of Entity Graph Extraction (EGX) in EgoLifeQA in ~\cref{tab:egolife_bm25_vs_llm_search}. The additional information from visual captions increases MCQ accuracy by $\sim$2.6\% on average across all three MLLM backbones (GPT-4o, GPT-4.1 and Gemini 2.5 Pro).

\noindent
\textbf{Agent Wall-Clock Latency.}
We tabulate the wall-clock latency of \mdlname{} pipeline in \cref{tab:quality_vs_latency}. \mdlname{} takes between two and three minutes to answer an MCQ, depending on the number of sub-tasks required by the planning agent. We also evaluate the latency impact of the transcript search and replace the default LLM search with BM25~\citep{robertson2009bm25}, which drops token usage by $3.3\times$ at the cost of a $\sim$6.8\% MCQ accuracy drop on average.

We discuss more ablations on tool usage and retrieval quality in \Cref{app:egolife}.
\section{Conclusion}
\label{sec:conclusion}
We introduce a novel \mdlname{} framework (\cref{ssec:agentic_framework}) for longitudinal video understanding, addressing the unique challenges posed by always-on personal AI assistants processing very long egocentric video streams. By leveraging entity scene graphs (\cref{ssec:eg_rep}) and specialized tools for structured, cross-modal reasoning, our approach enables detailed and temporally coherent analysis. Experiments on EgoLifeQA (\cref{ssec:experiments_egolife}) and Video-MME (Long) (\cref{ssec:experiments_videomme}) demonstrate state-of-the-art performance on tasks requiring the tracking of entities, behaviors, and relationships over extended periods. As video lengths continue to grow, we believe our results highlight the potential of agentic planning over structured representations of inter-entity relationships for very long video understanding moving forward.

\section{Limitations}
\label{sec:limitations}
While our \mdlname{} achieves strong performance on longitudinal video understanding tasks, it is important to note that the construction of entity scene graphs depends on the accuracy of upstream perception and language models, which may occasionally introduce errors in extracting entities and relationships. Additionally, our experiments relied on transcripts and for EgoLife, manually annotated speaker diarization. In scenarios where off-the-shelf diarization models are used, downstream performance is likely to be adversely affected by prediction errors. For specific examples of failure cases, please refer to \cref{app:failure_cases}.

\section{Ethical Considerations}
\label{sec:ethical}
Our work uses the publicly available EgoLife dataset, which was released under an MIT license. We adhere to all terms of use associated with this dataset. The EgoLife dataset automatically detects and blurs faces and other personally identifiable information (PII) such as sensitive audio content. We also use the Video-MME dataset, which was released under a custom license\footnote{License: \url{https://github.com/MME-Benchmarks/Video-MME}}. We have adhered to all terms of use associated with this dataset, using an unmodified version strictly for academic research. In addition to these pre-existing safeguards, we have taken extra care to protect individual privacy in our reporting: all faces appearing in the figures throughout this paper have been manually blurred.

\section*{Acknowledgements}
This work was supported in part by NSF IIS2404180 and Institute of Information \& communications Technology Planning \& Evaluation (IITP) grant funded by the Korea government (MSIT) No. 2022-0-00871 (Development of AI Autonomy and Knowledge Enhancement for AI Agent Collaboration) and No. RS-2025-2543949 (Environment-Aware and Domain-Adaptive Multimodal Embodied AI for RealWorld Interaction).

\bibliographystyle{assets/plainnat}
\bibliography{main}

\clearpage
\beginappendix

\setcounter{figure}{4}
\setcounter{table}{4}

\section{Overview}
\noindent\textbf{Design Details and Qualitative Examples.} We provide details of \mdlname{} design and a visual walkthrough of our entire \mdlname{} pipeline in \cref{app:agent_design} with a qualitative example. We demonstrate how the planning agent invokes retrieval tools to retrieve relevant context from the very long video and continuously update the working memory. We illustrate how we query our entity graph in \cref{app:eg_extraction}.

\noindent\textbf{Ablations on EgoLifeQA.}
We provide additional empirical analyses on EgoLifeQA~\citep{yang2025egolife} in \cref{app:egolife}, including evaluating oracle search, the importance of each search tool, retrieval accuracy of our three search tools, and wall-clock latency and memory cost of each component of \mdlname{}.

\noindent\textbf{Implementation Details.}
We provide the prompts and code snippets we use for our planning agent, to extract and temporally annotate our entity graph, to query our search tools, and other implementation details in \cref{app:implementation}.

\section{Qualitative Example of EGAgent Pipeline}
\label{app:agent_design}

\begin{figure*}[!t]
\centering
\includegraphics[width=.95\linewidth]{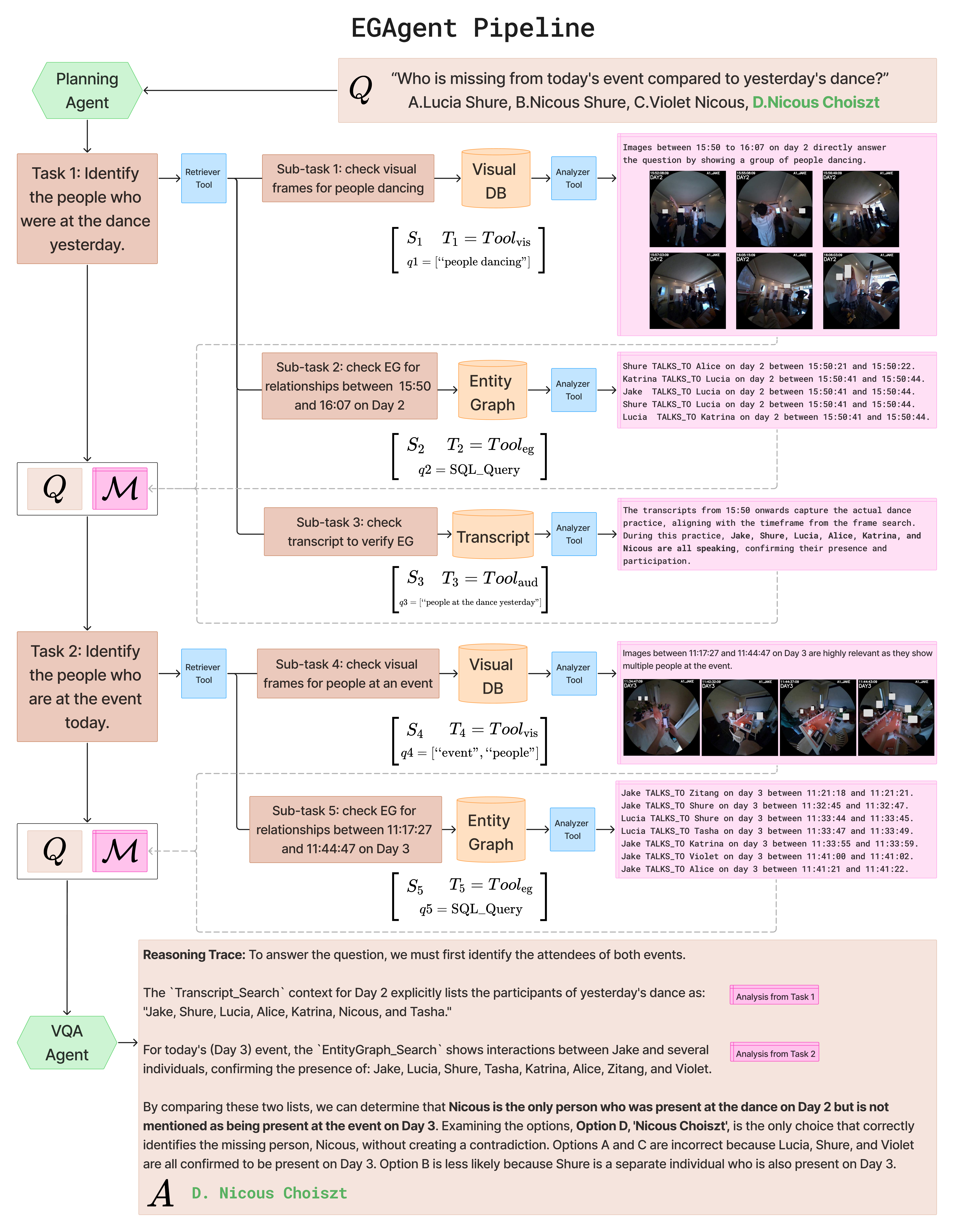}
\caption{A walkthrough of our entire \mdlname{} pipeline (Sec 3.3, main paper) for an example query from EgoLifeQA, with more details in \cref{app:agent_design}. 
At a high-level, given the \textcolor{query}{\textbf{query}}, the \textcolor{agent}{planning agent} comes up with a sequence of 5 sub-tasks, \emph{i.e.}, $S_1$ through $S_5$. Each sub-task is routed to one or more search tools $T_i$, followed by the \textcolor{retrievertool}{analyzer tool}, whose output is appended to the \textcolor{workingmemory}{working memory} $\mathcal{M} \gets\mathcal{M} \cup \text{Analysis}$. Once all sub-tasks are complete, the original query $Q$ and working memory $\mathcal{M}$ are sent to the VQA agent to predict the answer $A$. The \text{SQL\_Query} and the details about the entity graph search is illustrated in \cref{fig:app_full_pipeline_eg_querying}.}
\vspace{-4mm}
\label{fig:app_full_pipeline}
\end{figure*}

\begin{figure*}[!t]
\centering
\includegraphics[width=\linewidth]{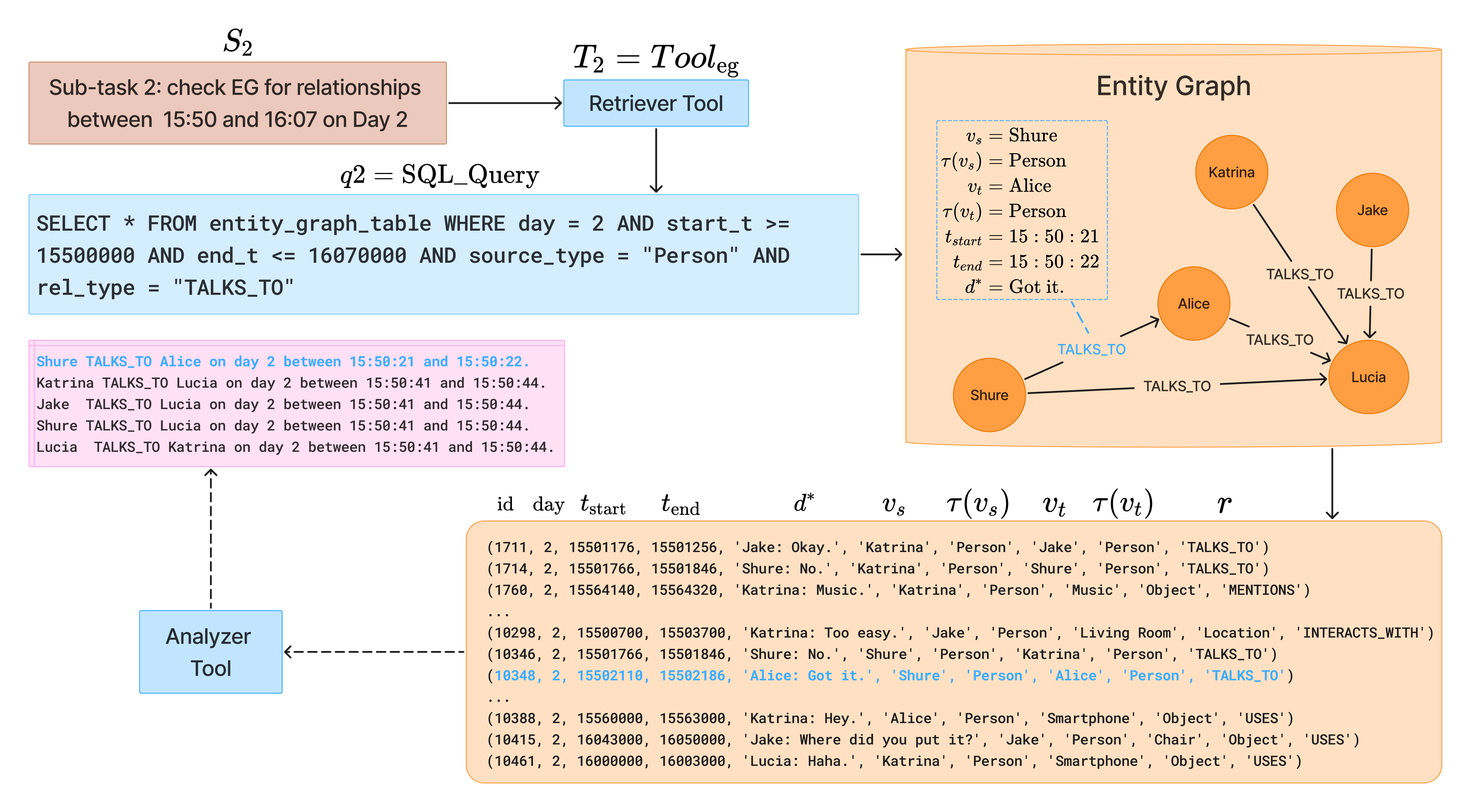}
\caption{Here we focus on the \textcolor{database}{entity graph} search tool $Tool_\mathrm{eg}$ in the example from \cref{fig:app_full_pipeline} and discuss its role in the overall \mdlname{} pipeline in \cref{app:agent_design}. Given the sub-task $S_2$, the \textcolor{agent}{planning agent} uses a strict-to-relaxed hierarchy to choose a SQL query $q_2$ to search the entity graph to answer the sub-task, \emph{i.e.}, graph entities $\tau(v_s) = \text{Person}$, $r=\text{TALKS\_TO}$ and $(\text{day}, t_{\text{start}},t_{\text{end}})$ to search between. The relevant rows of the SQL table, i.e., the relevant inter-entity relationships $(v_s, \tau(v_s), v_t, \tau(v_t), r, t_\mathrm{start}, t_\mathrm{end}, d^*)$ are appended to the \textcolor{workingmemory}{working memory} $\mathcal{M}$.}
\label{fig:app_full_pipeline_eg_querying}
\end{figure*}

We illustrate an example of our entire pipeline on a query from EgoLifeQA in \cref{fig:app_full_pipeline}. Given the \textcolor{query}{\textbf{query}}, the \textcolor{agent}{planning agent} identifies high-level tasks, and comes up with a sequence of $N$ sub-tasks ($N<6$). In this example, the planner generated 5 tasks, $S_1$ through $S_5$. Each planner step is routed to one or more search tools via route\_plan. When multiple modalities are needed for the same logical sub-task, the planner may repeat the same sub-task description across consecutive steps, each with a different $T_i$ (see \cref{ssec:agentic_framework}). In this example, $S_1$ is routed to $Tool_\mathrm{vis}$ to select relevant frames from the \textcolor{database}{Visual DB} with query $q_1 = \text{``people dancing''}$. These retrieved frames are then sent to the \textcolor{retrievertool}{analyzer tool}, which observes that people are dancing on day 2 between 15:50 and 16:07, without knowledge of their identities. Similarly, $S_2$ is routed to $Tool_\mathrm{eg}$ to search for social relationships in \textcolor{database}{Entity Graph}, which we describe in more detail in \cref{fig:app_full_pipeline_eg_querying}. Given the sub-task $S_2$, the \textcolor{agent}{planning agent} uses a strict-to-relaxed hierarchy to choose a SQL query $q_2$ to search the entity graph to answer the sub-task, \emph{i.e.} graph entities $\tau(v_s) = \text{Person}$, $r=\text{TALKS\_TO}$ and $( t_{\text{start}},t_{\text{end}})$ to search between. The retrieved rows of the SQL table, i.e., the relevant inter-entity relationships $(v_s, \tau(v_s), v_t, \tau(v_t), r, t_\mathrm{start}, t_\mathrm{end}, d^*)$ are appended to the \textcolor{workingmemory}{working memory} $\mathcal{M}$ (no separate analyzer call for entity-graph retrieval). We \textcolor{retrievertool}{highlight} one such relationship in \cref{fig:app_full_pipeline_eg_querying}, \emph{i.e.}, Shure saying ``Got it.'' to Alice between 3:50:21 PM and 3:50:22 PM, which overlaps with the dancing activity ($S_1$ in \cref{fig:app_full_pipeline}), indicating that both Shure and Alice take part in dancing. The planning agent proceeds until all remaining sub-tasks are routed to their appropriate search tool $T_i$ with query arguments $q_i$ and analyzed by the analyzer tool. The analysis output from each subsequent tool $S_3, S_4, S_5$ is also appended to the \textcolor{workingmemory}{working memory} $\mathcal{M}$. Once all sub-tasks are complete, the original query $Q$ and working memory $\mathcal{M}$ are sent to the VQA agent to predict the answer $A$. 

\begin{figure}[t]
    \centering
    \includegraphics[width=.8\linewidth]{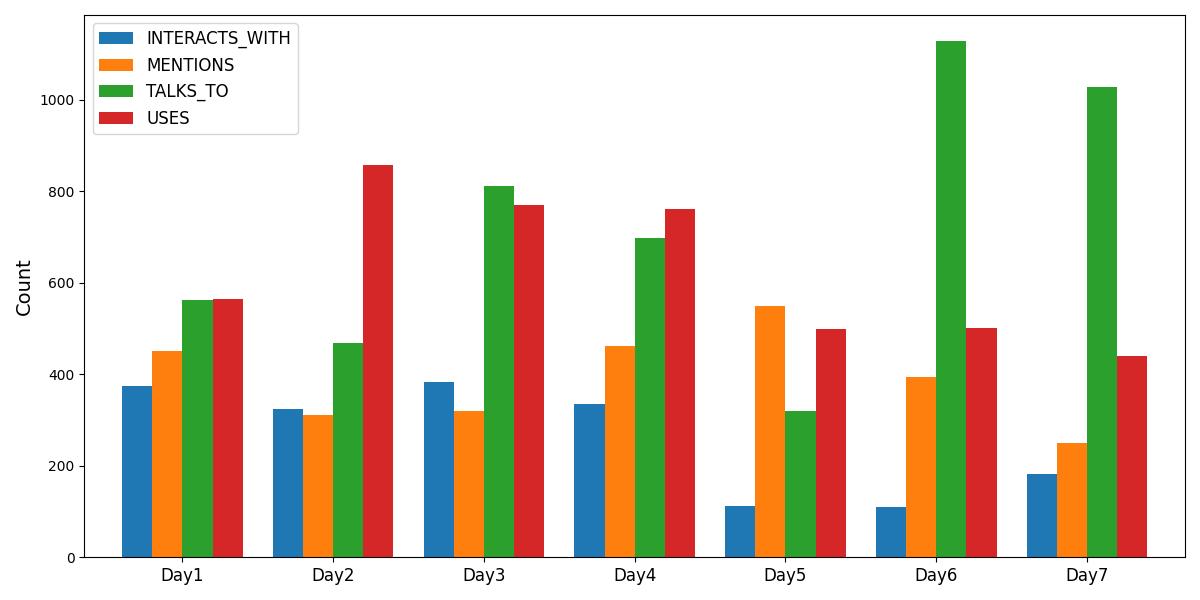}
    \caption{Entity Graph relationship types extracted from all seven days of EgoLife.}
    \label{fig:egolife_rels}
\end{figure}

\section{Entity Graph}
\label{app:eg_extraction}

We show a qualitative example of how we query the entity graph in our \mdlname{} pipeline in \cref{fig:app_full_pipeline_eg_querying}. We also discuss our temporal annotation of entity graph edges, a novel contribution that enables \mdlname{} to temporally localize relevant relationships for a given query. We provide the implementation details and the prompts we use to construct the entity graph and temporally annotate edges in \cref{app:implementation}. 

We also provide some statistics of the entity graph we extract from EgoLife. In total, we extract 13968 relationships over a 7 day period. We visualize the relationships extracted for each day in~\cref{fig:egolife_rels}. A vast majority of relationships have source node ``person'' (13930 / 13968), while the target node is more balanced (1314 ``location'', 6449 ``object'' and 6167 ``person''). This indicates that the graph is focused more on person-person and person-object interactions, while also capturing person-location information.

\section{Ablation Study on EgoLife}
\label{app:egolife}
Here we provide some additional ablation studies on EgoLife. We focus on tool usage, upper bound performance when using oracles, retrieval accuracy of tool search, and wall clock latency of our \mdlname{} pipeline. 

\begin{table*}[!t]
\centering
\caption{Ablation study on impact of each tool on MCQ accuracy across EgoLifeQA task types. We equip \mdlname{} with various combinations of search tools (EG for $Tool_\mathrm{eg}$, F for $Tool_\mathrm{vis}$, and T for $Tool_\mathrm{aud}$). \mdlname{} highlights the importance of cross-modal reasoning (EG, F, T) by showing strong performance on all task types, especially those requiring inter-entity relationships (RelationMap).
}
\resizebox{\textwidth}{!}{%
\begin{tabular}{@{}l|c|cccccc|cc@{}}
\toprule
\multicolumn{1}{c|}{\multirow{2}{*}{\textbf{Method}}} & \multirow{2}{*}{\textbf{Modality}} & \multicolumn{6}{c|}{\textbf{MCQ Acc (\%)}}                                                                                                              & \multirow{2}{*}{\textbf{\begin{tabular}[c]{@{}c@{}}Average\\ Gain (\%)\end{tabular}}} & \multirow{2}{*}{\textbf{\begin{tabular}[c]{@{}c@{}}Average\\ \# Tokens\end{tabular}}} \\ \cmidrule(lr){3-8}
\multicolumn{1}{c|}{}                                 &                                    & \textbf{EntityLog} & \textbf{EventRecall} & \textbf{HabitInsight} & \textbf{RelationMap} & \textbf{TaskMaster} & \multicolumn{1}{c|}{\textbf{Average}} &                                                                                       &                                                                                       \\ \midrule
EgoButler Gemini 1.5 Pro                             & C, T                               & 36.0               & 37.3                 & 45.9                  & 30.4                 & 34.9                & 36.9                                  & +0                                                                                    & --                                                                                    \\ \midrule
\mdlname{} GPT-4.1 (EG)                              & C                                  & 38.4               & 42.9                 & 31.1                  & 31.2                 & 44.4                & 37.6                                  & +0.7                                                                                  & 21K                                                                                   \\
\mdlname{} GPT-4.1 (F)                               & F                                  & 40.0               & 37.3                 & 31.1                  & 28.0                 & 34.9                & 34.6                                  & -2.3                                                                                  & 131K                                                                                  \\
\mdlname{} GPT-4.1 (T)                               & T                                  & 32.8               & 42.9                 & \textbf{59.0}         & 44.0                 & 66.6       & 45.6                                  & +8.7                                                                                  & 438K                                                                                  \\
\mdlname{} GPT-4.1 (F + T)                           & F, T                               & \textbf{48.0}      & 48.4                 & 55.7                  & 40.0                 & 61.9                & 48.6                                  & +11.7                                                                                 & 560K                                                                                  \\
\mdlname{} GPT-4.1 (EG+ F + T)                       & F, C, T                            & 44.0               & \textbf{49.2}        & 55.7                  & \textbf{53.6}        & \textbf{66.7}                & \textbf{50.7}                         & +13.8                                                                                 & 571K                                                                                  \\ \bottomrule
\end{tabular}
}

\label{tab:app_tool_ablation}
\end{table*}

\subsection{Ablation on tool usage}
\label{app:tool_use_ablation}

To evaluate the importance of each search tool $T$ on \mdlname{} performance, we evaluate our \mdlname{} with all possible combinations of tools in \cref{tab:app_tool_ablation}.

We observe that using only the frame search tool performs poorly as the agent has no sense of entity identities. This reflects in its near-random performance on RelationMap (28\%), while its performance on more visual-focused tasks like EntityLog (40\%) and EventRecall (37.3\%) remains strong compared to EgoButler. When we add the powerful audio transcript search tool to \mdlname{} (T), the accuracy significantly improves for HabitInsight (+13.1\%), RelationMap (+13.6\%), and TaskMaster (+31.7\%), while dropping slightly on the more visual-focused EntityLog (-3.2\%). Using only the audio transcript search tool $Tool_\mathrm{aud}$, \mdlname{} (T) performs the best on HabitInsight and TaskMaster, as these types of questions are more dependent on repeated and time-localized utterances from the audio transcripts. When we add the visual search tool $Tool_\mathrm{vis}$ to \mdlname{} (F+T), it slightly drops performance on audio and entity-relationship focused tasks HabitInsight (-3.3\%), TaskMaster (-4\%), and RelationMap (-5\%) compared to \mdlname{} (T), but similar to \mdlname{} (F), improves significantly on visual-focused tasks EntityLog (+15.2\%) and slightly on EventRecall (+5.5\%). Finally, when we add the entity graph search tool $Tool_\mathrm{eg}$, we get state-of-the-art performance on entity-focused RelationMap (+23.2\% over EgoButler), TaskMaster (+31.8\% over EgoButler) and EventRecall (+11.9\% over EgoButler), while remaining competitive on EntityLog and HabitInsight.

In summary, equipping \mdlname{} with the entity graph search tool $Tool_\mathrm{eg}$ in addition to the standard visual and audio search tools $Tool_\mathrm{vis}$ and $Tool_\mathrm{aud}$ is crucial for robust performance on tasks requiring knowledge distributed across modalities, \emph{e.g.}, inter-entity relationships (RelationMap, HabitInsight),  audio triggers (TaskMaster), and visual-focused tasks (EntityLog, EventRecall). This result indicates that for agents to robustly understand long videos, it is important that they can search across modalities and reason over this fused context (cross-modal reasoning).

\subsection{Oracles Indicate Room for Growth in Temporal Localization}
\label{app:oracle_accuracy}

\begin{table}[!b]
\centering
\caption{We use the ground-truth timestamps provided by EgoLifeQA to evaluate visual and audio transcript oracles, \emph{i.e.}, search has perfect precision (1.0).
F = raw video frames, C = video captions, A = raw audio, T = audio transcript. ``1 FPS~$\rightarrow$~50'' denotes retrieving 50 frames from those sampled at 1 FPS, with only these 50 frames used for MLLM analysis. We observe that there is still a gap between \mdlname{} tool search and perfect search, but perfect search still saturates at sub 70\% accuracy with the state-of-the-art multimodal LLM. }
\resizebox{.7\columnwidth}{!}{%
    \begin{tabular}{@{}l|cc|cc|c@{}}
    \toprule
    \multicolumn{1}{c|}{\multirow{2}{*}{\textbf{Method}}} & \multirow{2}{*}{\textbf{\# Frames}} & \multirow{2}{*}{\textbf{Modality}} & \multirow{2}{*}{\textbf{\begin{tabular}[c]{@{}c@{}}MCQ \\ Acc (\%)\end{tabular}}} & \multirow{2}{*}{\textbf{\begin{tabular}[c]{@{}c@{}}Average\\ Gain (\%)\end{tabular}}} & \multirow{2}{*}{\textbf{\begin{tabular}[c]{@{}c@{}}Average\\ \# Tokens\end{tabular}}} \\
    \multicolumn{1}{c|}{}                                 &                                     &                                    &                                                                                   &                                                                                       &                                                                                       \\ \midrule
    EgoButler Gemini 1.5 Pro                             & 0                                   & C, T                               & 36.9                                                                              & +0                                                                                    & -                                                                                     \\ \midrule
    GPT 4.1 Prev4Day                                     & 0                                   & T                                  & 45.6                                                                              & +8.7                                                                                  & 700K                                                                                  \\ \midrule
    \multirow{2}{*}{GPT 4.1 Oracle}                      & 0                                   & T                                  & 52.0                                                                              & +15.1                                                                                 & 243K                                                                                  \\
    & 50                                  & F, T                               & 57.6                                                                              & +20.7                                                                                 & 274K                                                                                  \\ \midrule
    \multirow{2}{*}{Gemini 2.5 Pro Oracle}               & 0                                   & T                                  & 57.9                                                                              & +21.0                                                                                 & 332K                                                                                  \\
    & 50                                  & F, T                               & \textbf{68.7}                                                                     & \textbf{+31.8}                                                                                & 346K                                                                                  \\ \midrule
    \mdlname{} GPT-4.1 (EG+F+T)                     & 1FPS$\to$50                         & F, C, T                            & 50.7                                                                              & +13.8                                                                                 & 571K                                                                                  \\
    \mdlname{} Gemini 2.5 Pro (EG+F+T)             & 1FPS$\to$50                         & F, C, T                            & 57.5                                                                              & +20.6                                                                                 & 880K                                                                                  \\ \bottomrule
    \end{tabular}
}
\label{tab:app_oracle_egolife}
\end{table}

To evaluate the upper bound performance, we use the ground-truth relevant moments (target\_time) as oracle information for visual and audio transcript search. For visual search, we uniformly sample 50 frames at 1 FPS centered on the timestamps from (target\_time), and for audio transcript search, we extract the entire transcript from the ground-truth day. As seen in \cref{tab:app_oracle_egolife}, using both a visual and audio transcript oracle outperforms \mdlname{} (EG + F + T) by $6.9\%$ with GPT 4.1 and $11.2\%$ with Gemini 2.5 Pro. This shows that there is still room for improvements in MCQ accuracy that can be enabled by better temporal localization over very long videos.

\subsection{Retrieval Accuracy}
\label{app:retrieval_accuracy}
\begin{table*}[t!]
\centering
\caption{Recall@windowsize (recall@W) of agentic approaches on EgoLifeQA with respect to ground-truth timestamps provided by the dataset. We compute recall over temporal windows W centered on each ground-truth timestamp for the predicted timestamps from each tool: \emph{i.e.}, EG for $Tool_\mathrm{eg}$, F for $Tool_\mathrm{vis}$, and T for $Tool_{aud}$. The number of timestamps that each search tool searches over is marked by Input \#ts, and the number of timestamps highlighted by the analyzer tool as relevant to the query is marked by Selected \#ts.}
\resizebox{.9\textwidth}{!}{%
\begin{tabular}{@{}c|l|cc|cccccc@{}}
\toprule
\multirow{2}{*}{\textbf{Category}}                                 & \multicolumn{1}{c|}{\multirow{2}{*}{\textbf{Method}}} & \multirow{2}{*}{\textbf{\begin{tabular}[c]{@{}c@{}}Input \\ \# ts\end{tabular}}} & \multirow{2}{*}{\textbf{\begin{tabular}[c]{@{}c@{}}Selected \\ \# ts\end{tabular}}} & \multicolumn{6}{c}{\textbf{recall@W}}                                                                   \\ \cmidrule(l){5-10} 
& \multicolumn{1}{c|}{}                                 &                                                                                 &                                                                                    & \textbf{10 sec} & \textbf{30 sec} & \textbf{1 min} & \textbf{2 min} & \textbf{10 min} & \textbf{1 hour} \\ \midrule
\begin{tabular}[c]{@{}c@{}}MLLMs\\ (Uniform Sampling)\end{tabular} & Gemini 2.5 Pro                                       & 3000                                                                            & ~~3.1                                                                               & 0.101           & 0.160           & 0.192          & 0.238          & 0.325           & 0.410           \\ \midrule
\multirow{5}{*}{Ours}                                              & \mdlname{} (F+T) Overall                             & 4750                                                                                & ~~4.8                                                                               & 0.232           & 0.241           & 0.255          & 0.268          & 0.322           & 0.418           \\ \cmidrule{2-10}
& \mdlname{} (EG+F+T) $\mathcal{M}_{EG}$               & ~~158                                                                             & 10.8                                                                               & 0.127           & 0.166           & 0.199          & 0.233          & 0.413           & 0.658           \\
& \mdlname{} (EG+F+T) $\mathcal{M}_{VIS}$              & ~~~~50                                                                              & 17.6                                                                               & 0.857           & 0.868           & 0.873          & 0.875          & 0.900           & 0.930           \\
& \mdlname{} (EG+F+T) $\mathcal{M}_{AUD}$              &  4700                                                                               & ~~2.6                                                                               & 0.218           & 0.247           & 0.261          & 0.288          & 0.347           & 0.417           \\
& \mdlname{} (EG+F+T) Overall                          & 4958                                                                                & 31.0                                                                                 & 0.884           & 0.895           & 0.898          & 0.902          & 0.932           & 0.962           \\ \bottomrule
\end{tabular}
}
\label{tab:app_retrieval_acc}
\end{table*}

Our oracle upper bound experiments in \cref{app:oracle_accuracy} highlight that precise temporal localization enables strong MCQ accuracy on EgoLifeQA, and that retrieval quality is an important factor in the success of agentic approaches for very long video understanding. To evaluate where the strength of our agent is coming from, we do a simple recall analysis on EgoLifeQA. We examine the working memory $\mathcal{M}$ for each multiple-choice question in the dataset, and extract the portions added by each search tool $\mathcal{M}_{eg}$ by $Tool_\mathrm{eg}$, $\mathcal{M}_{vis}$ by $Tool_\mathrm{vis}$, and $\mathcal{M}_{aud}$ by $Tool_\mathrm{aud}$.

Each multiple-choice question in EgoLifeQA $\text{mcq}_i$ contains $\text{total\_i}$ ground-truth timestamps in (target\_time). To evaluate the quality of search of our tools, we compute a recall over these ground-truth timestamps with each of our search tools in \cref{tab:app_retrieval_acc}. We denote the number of timestamps that each search tool searches over by ``Input \#ts'', and the number of relevant timestamps highlighted by the analyzer tool (which is added to working memory $\mathcal{M}$) by ``Selected \#ts''. For visual and BM25 transcript search, ``Selected \#ts'' comes from the analyzer output; for LLM transcript search, from the combined retrieval/relevance call; for entity-graph search, from timestamps in the formatted SQL rows. For example, the Multimodal LLM baseline (Gemini 2.5 Pro) is provided 3000 uniformly sampled timestamps (Input \#ts), from which it selects $\sim3.1$ as relevant to the query (Selected \#ts).

Since the provided target\_time are discrete (HH:MM:SS), we use time windows centered on the target\_time in our computation. For a given $\text{mcq}_i$ and search tool, we record a $\text{hit}_{w,i}$ if any timestamp selected by the search tool (\emph{i.e.}, one of Selected \#ts) lies in the temporal window W. We define recall@windowsize (recall@W) over the $N=500$ MCQ in EgoLifeQA as follows:

$$
recall@W = \sum\limits_{i=1}^{N} \dfrac{\text{hits}_{w, i}}{\text{total}_i}
$$

We vary the size of these windows from 10 seconds up to one hour to measure how recall saturates as we relax the strictness of temporal localization. As seen in \cref{tab:app_retrieval_acc}, the visual search tool shows strong recall even at window size of 10 seconds, indicating that it shows strong temporal localization capabilities. It is natural to question why our MCQ accuracy remains relatively low (34.6\% when using only $Tool_\mathrm{vis}$, \cref{tab:app_tool_ablation}) even with such high recall of $Tool_\mathrm{vis}$; we highlight that even with perfect precision (using an audio-visual oracle, \cref{app:oracle_accuracy}), the MCQ accuracy saturates at 68.7\%. This indicates that an audio-visual analysis of ground-truth timestamps alone is insufficient to push the frontier further.

The audio transcript search tool shows poor recall at small window sizes, which is surprising as an oracle with audio transcript search is 21\% better than the previous state-of-the-art (Gemini 2.5 Oracle with T in \cref{tab:app_oracle_egolife}). When examining $\mathcal{M}_{\text{aud}}$ we observe that this is because while the analyzer tool points out relevant context from audio transcripts for each task from the planning agent, it occasionally misses explicitly pointing out timestamps when the relevant timestamp is ambiguous. This leads to missing hits even when the analyzer has analyzed the correct portion of the audio transcript (as is evident from the search tool's analysis in the working memory $\mathcal{M}$). 

\begin{table*}[!ht]
\caption{An expanded version of \Cref{tab:quality_vs_latency} showing wall-clock latency in seconds of each module within \mdlname{} averaged over all MCQ on EgoLifeQA. For both the Visual and Transcript searches, the wall-clock time of the analyzer tool (a multimodal LLM) dominates the retrieval time. When the transcript search backbone is an LLM, both the retrieval and analysis happen simultaneously.}
\centering
\resizebox{\textwidth}{!}{%
    \begin{tabular}{@{}c|cc|cccccccc|c@{}}
\toprule
\multirow{3}{*}{Method}                                                                                        & \multirow{3}{*}{Acc(\%)} & \multirow{3}{*}{\begin{tabular}[c]{@{}c@{}}Transcript\\ Search\\ Backbone\end{tabular}} & \multicolumn{8}{c|}{Wall-Clock Runtime (sec)}                                                                                                                                                                                               & \multirow{3}{*}{\#Tokens} \\ \cmidrule(lr){4-11}
&                          &                                                                                         & \multirow{2}{*}{Planning} & \multicolumn{2}{c}{Visual Search}                            & \multirow{2}{*}{EG Search} & \multicolumn{2}{c}{Transcript Search}                        & \multirow{2}{*}{VQA Agent} & \multirow{2}{*}{Total} &                           \\ \cmidrule{5-6} \cmidrule{8-9}
&                          &                                                                                         &                           & \multicolumn{1}{l}{Retriever} & \multicolumn{1}{l}{Analyzer} &                            & \multicolumn{1}{l}{Retriever} & \multicolumn{1}{l}{Analyzer} &                            &                        &                           \\ \midrule
\multicolumn{1}{l}{\multirow{2}{*}{\begin{tabular}[c]{@{}l@{}}\mdlname{} GPT-4.1\\ (EG + F + T)\end{tabular}}} & 43.9                     & BM25                                                                                    & 3.1                       & 4.6                           & 41.1                         & ~~8.4                      & 1.7                           & ~~8.2                        & 6.9                        & 125                    & 172K                      \\
\multicolumn{1}{l}{}                                                                                           & 50.7                     & LLM                                                                                     & 3.1                       & 4.5                           & 41.8                         & 10.2                       & --                           & 35.4                         & 6.9                        & 169                    & 571K                      \\ \bottomrule 
\end{tabular}
}
\label{tab:app_expanded_latency}
\end{table*}

The entity graph search tool shows the worst fine-grained temporal localization of all \mdlname{} search tools at smaller window sizes ($\leq$ 2 min) which is expected as it is a lower-dimensional projection of the audio-visual space when compared to visual embeddings in $\mathbb{R}^d$ generated by a vision encoder (SigLIP 2) or raw audio transcripts. We observe that the entity graph starts to beat the recall@W of the audio transcript search at windows $>$ 2 minutes, indicating its broader temporal coverage compared to the audio transcript search. Since searching the entity graph is $3.5\times$ faster than audio transcript search (\cref{tab:app_expanded_latency}), the entity graph search provides a flexible recall-latency tradeoff and is valuable to our \mdlname{} for both coarse temporal shortlisting (high recall at large window size) and fine-grained cross-modal reasoning with the visual and audio transcript search tools. See \Cref{fig:teaser} and \cref{fig:app_full_pipeline} for examples.

Lastly, when all our tools are combined to form \mdlname{} (EG + F + T), we observe very strong recall of 0.88 even at a window size of 10 seconds. This result provides evidence that the strong performance of \mdlname{} on EgoLifeQA (\Cref{tab:egolifeqa_results}) can be attributed to higher quality temporal localization of context relevant to the original query about the very long video.

\subsection{\mdlname{} Latency and Memory}
\label{app:Latency}
\noindent\textbf{Latency.} We expand Tab. 4 (main paper) to show latency of each component of \mdlname{} in \cref{tab:app_expanded_latency} in terms of wall clock time. We observe that the latency of the MLLM analyzer tool dominates for both the visual search and audio transcript search
compared to the actual retrieval time. Notably, in the case of visual search, the analyzer MLLM must process 50 retrieved frames, contributing significantly to the latency ($\sim9.1\times$ higher than the actual retrieval time). When switching the backbone of audio transcript retrieval from a MLLM to BM25, the latency of overall audio transcript search is $3.6\times$ lower. This analysis shows that our entity graph search tool adds minimal inference overhead to standard audio-visual search setups ($12.8\%$ on average) while providing strong accuracy gains, especially in tasks requiring knowledge of inter-entity relationships (\cref{tab:app_tool_ablation}).

\noindent\textbf{Memory.} The entity graph queried by \mdlname{}'s \textbf{Entity Graph Search Tool} ($Tool_\mathrm{eg}$) is extremely lightweight. For $\sim$52 hours of EgoLife video recorded by participant Jake, the SQLite database occupies only $\sim$2 MB on disk, and for Video-MME (Long), $\sim$65 KB on average per video. On the other hand, the visual embedding database used by the \textbf{Visual Search Tool} ($Tool_\mathrm{vis}$) is much larger as it stores embeddings of video frames sampled at 1 frame-per-second (FPS). On Egolife (Jake), this corresponds to 187011 frames ($\sim$51.94 hours) and 1.4 GB memory on disk. On Video-MME (Long), this corresponds to a total of 739896 frames ($\sim$205.52 hours) and 5.7 GB memory on disk, or 19 MB per video. The \textbf{Audio Transcript Search Tool} ($Tool_\mathrm{aud}$) queries raw audio transcripts (.srt files provided by the original dataset creators), which total 5.2 MB memory on disk on EgoLife (Jake). Video-MME provides transcripts for only 744 / 900 total videos and 292 / 300 videos in the Long subset. The Long subset transcripts which \mdlname{} queries occupies a total of 24.2 MB or $\sim$83 KB per video.

\subsection{Entity Graph Noise}
\label{app:eg_noise_analysis}
To address possible ASR or extraction errors affecting the quality of our entity graph relationships, we utilize a strict-to-relaxed search strategy where the agent automatically broadens temporal windows to maximize recall if initial matches fail (\Cref{ssec:agentic_framework}). Additionally, we randomly sample 100 relationships from EgoLife (out of 13968) and manually audit if the relationship is correct by examining the raw video and audio transcripts. On this subset, we find a \textbf{94\% accuracy rate}, where the failures generally correspond to subtle errors in visual captioning or while fusing the captions and transcripts.

\section{Failure Cases}
\label{app:failure_cases}
\begin{figure}[t]
    \centering
    \centering
    \begin{subfigure}{0.5\textwidth}
        \centering
        \includegraphics[width=\linewidth]{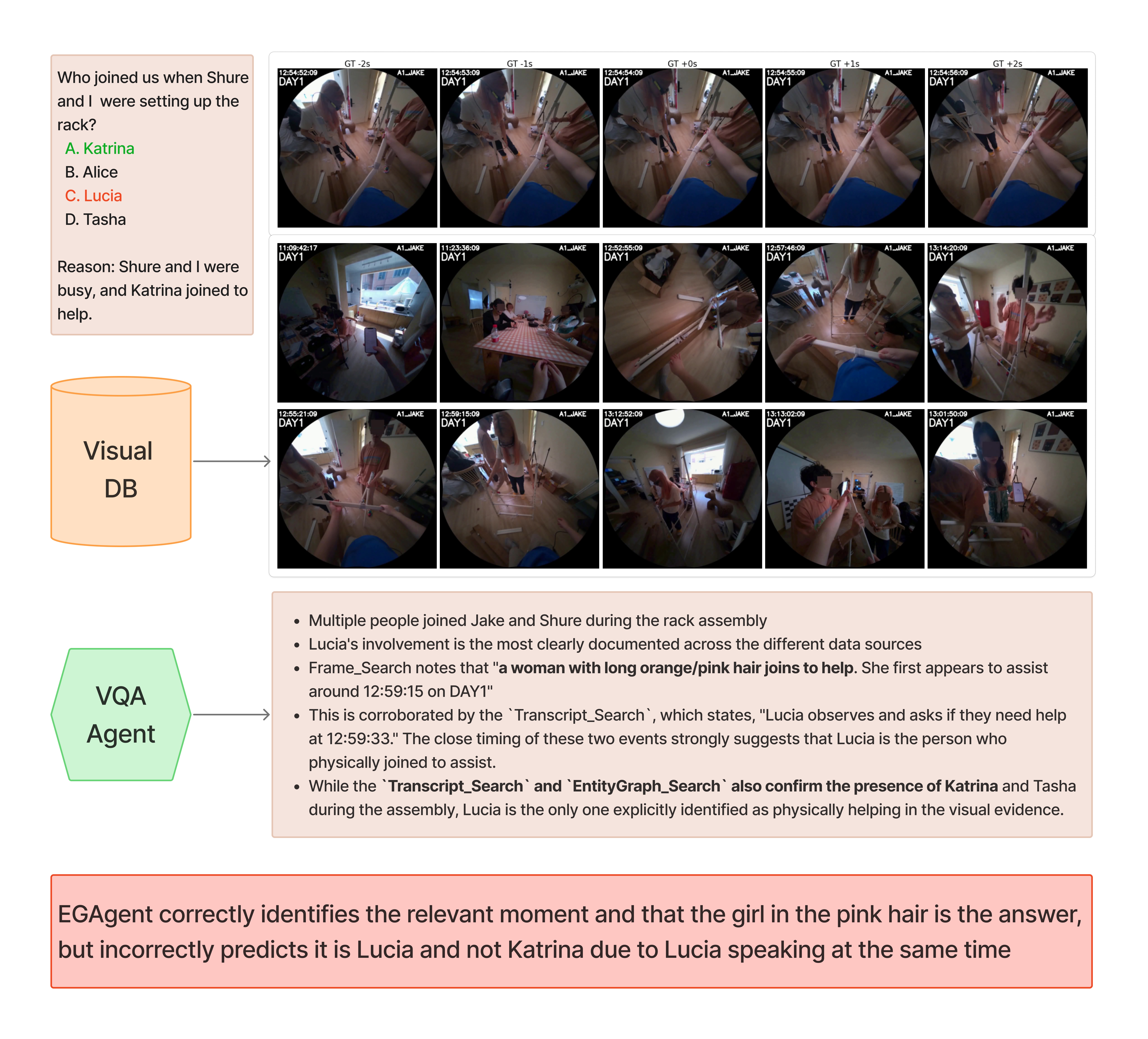}
        \caption{QID 18}
        \label{fig:mistakes1}
    \end{subfigure}%
    \begin{subfigure}{0.5\textwidth}
        \centering
        \includegraphics[width=\linewidth]{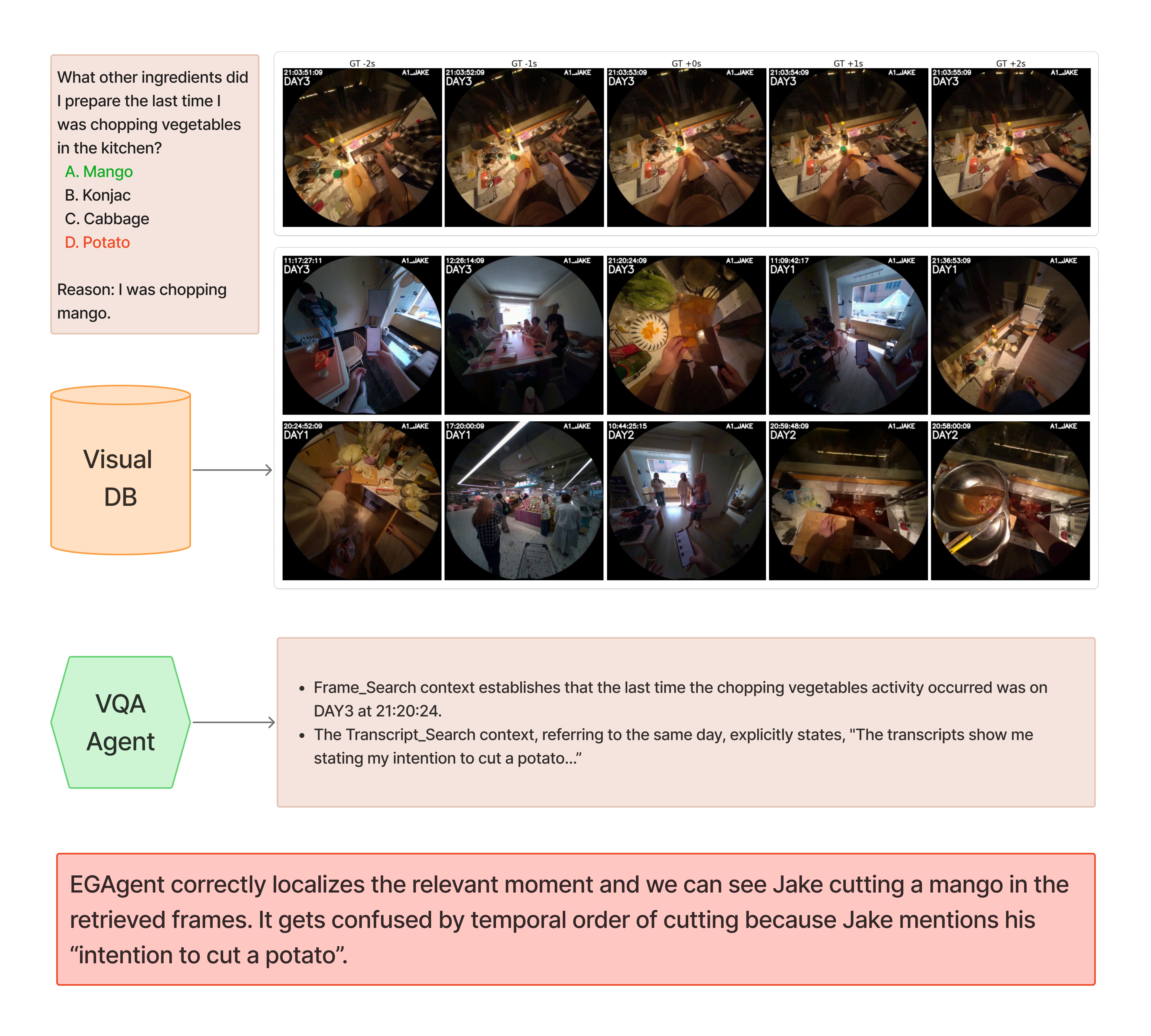}
        \caption{QID 231}
        \label{fig:mistakes2}
    \end{subfigure}
    \begin{subfigure}{0.48\textwidth}
        \centering
        \includegraphics[width=\linewidth]{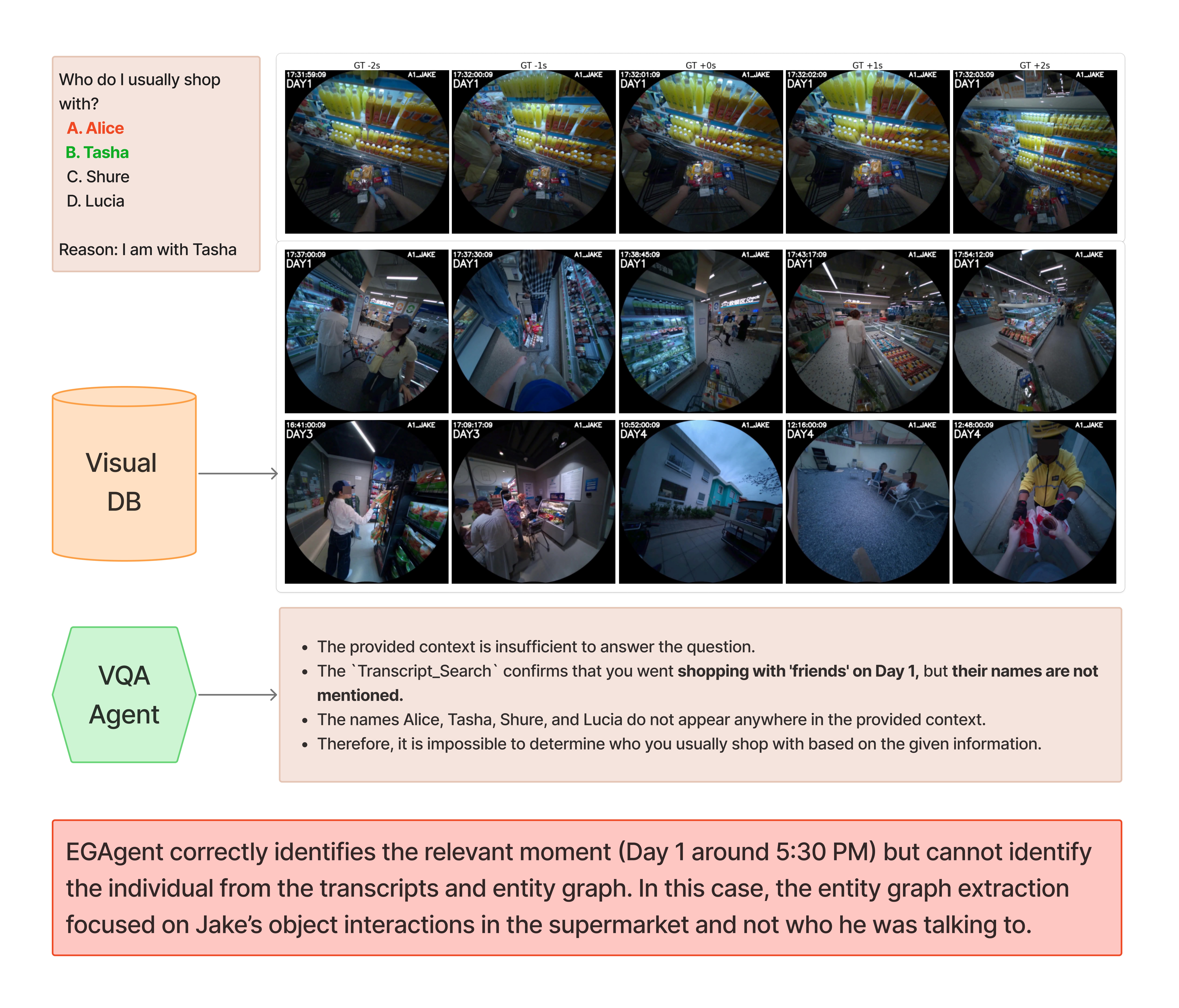}
        \caption{QID 353}
        \label{fig:mistakes3}
    \end{subfigure}
    \begin{subfigure}{0.48\textwidth}
        \centering
        \includegraphics[width=\linewidth]{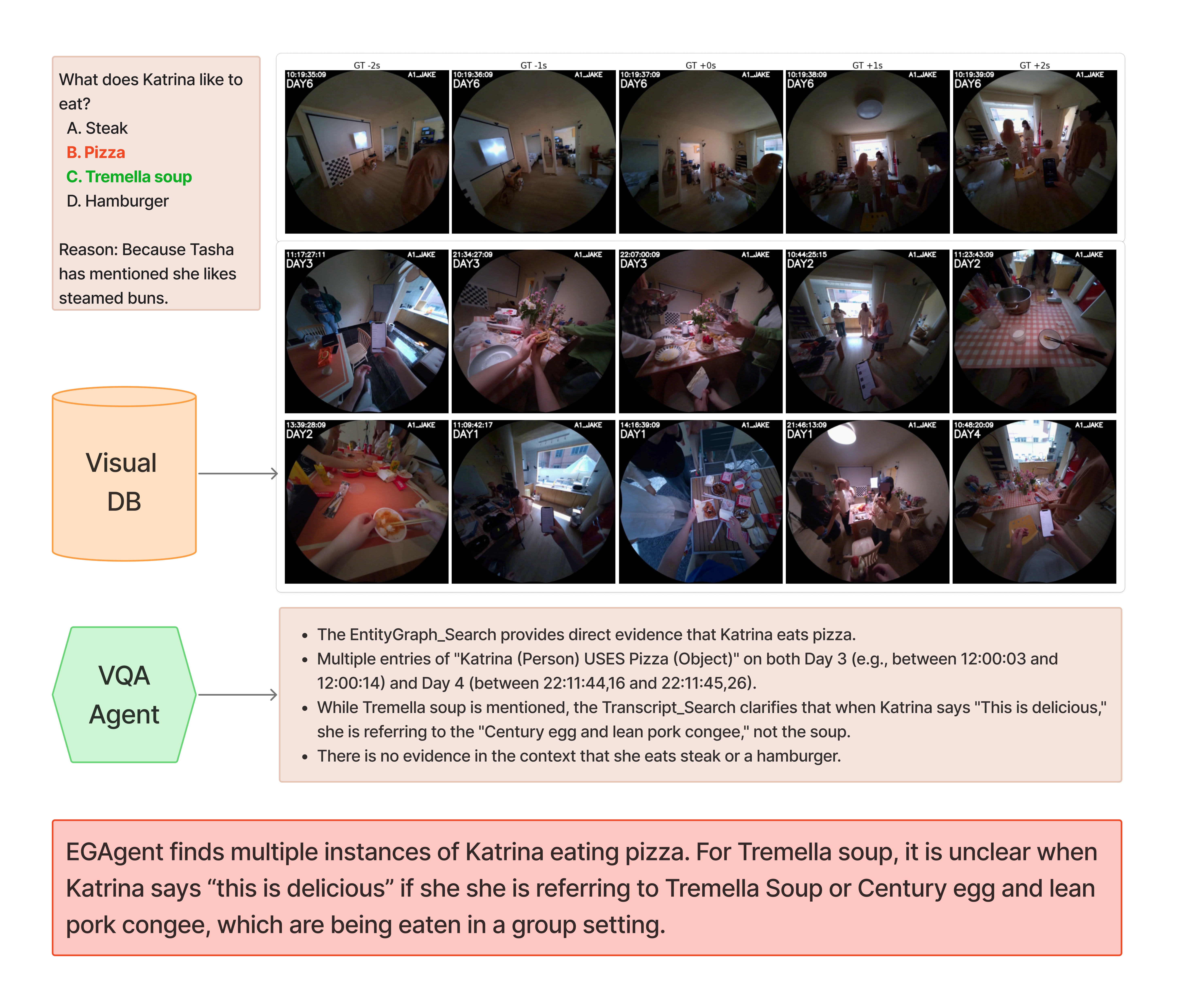}
        \caption{QID 470}
        \label{fig:mistakes4}
    \end{subfigure}
    \caption{Analyzing \mdlname{} mistakes on EgoLifeQA. The \textcolor{query}{query, options} are in the top left, with colors denoting the \textcolor{agent}{ground-truth answer} and the \textcolor{red}{agent's incorrect prediction}, as well as a ground-truth labeled ``reason'' for the correct answer. The topmost row shows a 4 second window around the \texttt{target\_time} for each query provided by the benchmark (ground-truth time for evidence). The two rows of images below the top show frames retrieved by the Visual Search Tool ($Tool_\mathrm{vis}$). The \textcolor{query}{box below the images} shows the reasoning chain of the \textcolor{agent}{VQA Agent} and the red box at the bottom summarizes \textcolor{red}{why \mdlname{} made a mistake}. QID for each example refers to the question ID from EgoLifeQA.}
    \label{fig:mistakes}
\end{figure}
We show example mistakes made by \mdlname{} (Gemini 2.5 Pro) on EgoLifeQA in \cref{fig:mistakes}. 

\noindent (1) \textbf{Identity Attribution under Perceptual Ambiguity.}
As shown in \cref{fig:mistakes1}, \mdlname{} successfully localizes the relevant moment but fails to reliably identify the individual in the frame. This highlights the challenges in multimodal identity grounding; the agent struggles to disambiguate individuals when visual signals are degraded and confounded by audio, misaligning identity with concurrent speech rather than visual context. Incorporating persistent persona representations (e.g., appearance cues like hair color or body shape) could improve robustness.

\noindent (2) \textbf{Conflicts Between Audio and Visual Evidence.}
\cref{fig:mistakes2} illustrates a failure case stemming from incorrect temporal reasoning due to conflicts between audio and visual evidence. \mdlname{} inconsistently reconciles temporal evidence across modalities, sometimes privileging prior verbal intent over more reliable visual observations. Enforcing cross-modal consistency checks and prioritizing direct evidence could mitigate this issue.

\noindent (3) \textbf{Entity Graph Incompleteness.}
In the example in \cref{fig:mistakes3}, social or relational information is underrepresented relative to object interactions in the graph, which \mdlname{} cannot reconcile. Richer representations of agents and interactions, along with more targeted query planning to explicitly target missing relational attributes, may address this limitation.

\noindent (4) \textbf{Ambiguity in the Question.}
As shown in \cref{fig:mistakes4}, when asked to infer Katrina's food preferences, \mdlname{} selects pizza, supported by multiple observations of her consuming pizza over the course of the week. However, during a shared meal, she remarks ``this is delicious" in the context of Tremella soup being discussed, introducing ambiguity as to whether the statement refers specifically to the soup or to another dish at the table. Both pizza (frequent consumption) and Tremella soup (positive verbal feedback) are plausible answers.
\clearpage

\section{Implementation Details}
\label{app:implementation}
For GPT-4.1, GPT-4o, and Gemini 2.5 Pro, we use the default settings with a temperature of 0 and a maximum of 3 retries. For Qwen-2.5-VL-7B (see Table 2 in the main paper), the model is hosted locally using vLLM on 4$\times$H200 GPUs, with temperature set to 0, \texttt{tensor-parallel-size} = 4, and \texttt{gpu-memory-utilization} = 0.85.

\noindent\textbf{Agent Implementation.}
We use LangGraph~\citep{LangGraph} for implementing our ~\mdlname{}. We use AI assistants to help write code for our agent implementation. We first convert our multiple-choice question into a StateGraph called VeryLongVideoQA which contains all necessary attributes for \mdlname{} inference. We show code for our agent design below. Note that all accuracies reported in this work are from a single run, as running agents multiple times on each dataset is computationally prohibitive.

We construct our \mdlname{} as shown in \Cref{fig:eg_creation} over the VeryLongVideoQA StateGraph. Once ~\mdlname{} receives a query $Q$, the planning agent (\texttt{planner}) decomposes $Q$ into a sequence of $N$ sub-tasks $\{S_1,\ldots,S_N\}$, stored in \texttt{VeryLongVideoQA.plan}.
For each sub-task $S_i$, the planner also selects one or more search tools $T_i$ (visual, entity graph, or audio transcript search), stored in \texttt{VeryLongVideoQA.plan\_steps}. The current sub-task $S_i$ being addressed by the agent is tracked in \texttt{VeryLongVideoQA.current\_task}.

A dispatch node (\texttt{route\_plan}) then executes the tools selected for $S_i$ in order.
For visual search and BM25 transcript search, each search node retrieves evidence and a separate analyzer LLM summarizes relevance to $S_i$. For LLM-based transcript search, retrieval and relevance summarization are performed in a single call. Entity-graph search appends formatted SQL results directly to $\mathcal{M}$. (\texttt{VeryLongVideoQA.working\_memory}). When all tools selected by the planner for $S_i$ have run, a \texttt{mark\_step\_complete} node records $S_i$ in \texttt{VeryLongVideoQA.previous\_tasks}. We also use an early exit condition that grades (\texttt{grade\_plan\_completion}) whether $\mathcal{M}$ already contains sufficient evidence for future sub-tasks to answer the full multiple-choice question (i.e., to discriminate among all four options). If it does not, we return control to the planning agent and proceed with the next sub-task $S_{i+1}$; otherwise, the agent early-exits to the VQA agent (\texttt{generate\_answer}), which predicts the final answer $A$ (\texttt{VeryLongVideoQA.answer}). If all $N$ sub-tasks have been executed without early exit, the agent also proceeds to \texttt{generate\_answer}.

\begin{python}
from typing_extensions import TypedDict
from typing import List, Literal
from pydantic import BaseModel, Field

# Planner output (per-step tool selection)
class PlanStep(BaseModel):
    task: str
    tools: List[Literal["visual", "audio", "eg"]]
    
class MultiHopPlan(BaseModel):
    steps: List[PlanStep]
    
# Defines graph attributes
class VeryLongVideoQA(TypedDict):
    question: str
    candidates: List[str]
    selected_video: str
    query_time: dict
    day_search_dict: dict
    audio_transcripts: list
    plan: List[str]                      # task strings (paper-facing)
    plan_steps: List[dict]               # {task, tools} per step
    working_memory: str
    current_task: str
    previous_tasks: List[str]
    remaining_tools_this_step: List[str] # tool queue for current step
    next_route: str                      # visual | audio | eg | step_done
    answer: str
    total_tokens: list
\end{python}
\clearpage

\begin{python}
from langgraph.graph import StateGraph, START, END

wf = StateGraph(VeryLongVideoQA)

# Define the agent nodes
wf.add_node("planner", planner_node)
wf.add_node("route_plan", route_next_tool_node)
wf.add_node("mark_step_complete", mark_step_complete_node)
wf.add_node("search_eg", search_entity_graph)
wf.add_node("search_visual", search_and_analyze_frames)
wf.add_node("search_transcripts", retrieve_transcripts)  # or BM25 variant
wf.add_node("generate_answer", generate_answer)

# Build agent graph
wf.add_edge(START, "planner")
wf.add_edge("planner", "route_plan")
# Dispatch tools selected for the current plan step
wf.add_conditional_edges(
    "route_plan",
    route_from_dispatch,
    {
        "visual": "search_visual",
        "eg": "search_eg",
        "audio": "search_transcripts",
        "step_done": "mark_step_complete",
    },
)

# Each tool updates working_memory, then returns to route_plan
wf.add_edge("search_visual", "route_plan")
wf.add_edge("search_eg", "route_plan")
wf.add_edge("search_transcripts", "route_plan")

# Allow early termination after all tools for the current step complete
wf.add_conditional_edges(
    "mark_step_complete",
    grade_plan_completion,
    {
        "complete": "generate_answer",
        "incomplete": "planner",
    },
)

# Send working memory and query to VQA Agent
wf.add_edge("generate_answer", END)
\end{python}

\noindent\textbf{Entity Graph Extraction.} To create an entity graph, we first need a good audio-visual scene representation to extract entities and relationships from. We create these scene representations by fusing (with GPT 4.1) audio transcripts and visual captions we generate via GPT-4.1 at 30 second intervals (see ``System Prompt for Visual Caption - Transcript Fusion'' below). These fused captions have cross-modal information, where people, objects, actions, and events are described by visual captions, and audio cues (+ speaker identities in the case of EgoLife) provide additional context to relationships that occur in the scene. 

We use Langchain's LLMGraphTransformer to extract an initial candidate set of nodes and relationships from our generated fused captions. While temporal localization via search tools is very important for long video understanding (\cref{app:oracle_accuracy}), LLMGraphTransformer module does not support adding any additional metadata to graph nodes and edges. To later equip our search tool with temporal filtering capabilities, we annotate all extracted relationships (entity graph edges) with timestamps based on the audio transcripts and visual captions. See ``User Prompt for Temporal Annotation of Entity Graph Edges'' below for more details.

\noindent\textbf{Entity Graph Extraction Design Choices.} We initially experimented with free-form relation extraction, allowing the LLM to describe relationships from the fused transcripts and video captions without constraints. We found nearly all of these relationships collapsed into four major buckets, i.e., ``interacts with'' and ``uses'' for physical person-object interactions in specific locations and ``talks to'' or ``mentions'' for verbal interactions, which are the abstractions we used to extract the entity graph (\texttt{allowed\_relationships} below). Increasing relation granularity primarily introduced synonymous or low-frequency relation types, which fragmented retrieval queries and drastically reduced entity-graph recall during agentic search (\cref{app:retrieval_accuracy}). We found that coarser abstractions improved robustness by enabling broader matching across noisy ASR and caption signals while preserving the relational information needed for temporal localization and multi-hop reasoning. We also highlight that while these four abstractions are sufficient for the dense social and physical interactions in egocentric data like EgoLife, our entity graph extraction pipeline is entirely domain-agnostic. The relationship schema itself is prompt-defined (see ``System Prompt for Entity Graph Extraction'' below), making it straightforward and efficient to adapt to different domains (e.g., 30 minutes on $\sim$52 hours of EgoLife).

\begin{python}
from langchain_core.documents import Document
from langchain_experimental.graph_transformers import LLMGraphTransformer

def generate_eg(text:str):
    llm = get_vision_llm('gpt-4.1')
    allowed_nodes = ["Person", "Location", "Object"]
    allowed_relationships = ["TALKS_TO", "INTERACTS_WITH", "MENTIONS", "USES"]
    docs = [Document(page_content=text)]
    eg = LLMGraphTransformer(
        llm,
        allowed_nodes,
        allowed_relationships
    )
    eg = eg.aconvert_to_graph_documents(docs)
    return eg

# fuse audio transcripts and visual captions
fused_captions = generate_fc(captions, transcripts)
relationships =  generate_eg(fused_captions)

eg_with_tstamp = temporal_annotator.
  invoke(
    {
        "relationships": relationships,
        "transcripts": transcripts, 
        "captions": fused_captions
    }
  )
\end{python}

\noindent\textbf{Token Usage Estimates.} Here we provide details for estimates of total tokens used by baseline methods in \cref{tab:egolifeqa_results} and \cref{tab:videomme_results}. For GPT 4.1 and Gemini 2.5 Pro, we apply 85 and 258 tokens per image respectively as per their API documentation. For Video-RAG \citep{luo2025videorag}, we add 2K tokens used by auxiliary texts (reported in the original paper) to an estimated 258 tokens per image. For EgoGPT \citep{yang2025egolife}, we roughly estimate the number of tokens for text summaries at intervals of 30 seconds ($\sim$100), one hour ($\sim$500), and one day ($\sim$2000). We assume one inference pass searches one day (2K tokens), 10 hours per day (5K tokens), and 120 30-second intervals per hour (12K tokens). For all other methods~\citep{xue2025adavideorag, ma2025drvideo, yuan2025videodeepresearch}, we assume they use the entire LLM context window.

\begin{tcolorbox}[colback=green!5, colframe=green!75!black, boxrule=1pt, arc=1.5pt, width=\textwidth ]
\rmfamily
\textbf{Planning Agent Prompt:}\vspace{0.3cm}

``````
You are an expert at answering questions about very long videos. These questions may require multi-hop reasoning. Your job is to come up with a multi-step plan of all possible information that may be needed to answer the question. 
\vspace{0.1cm}

Each step of your plan will be routed to one of three search tools. The first search tool looks at transcripts with timestamps, the second looks at image frames sampled at 1 FPS, and the third looks at an entity scene graph extracted from the long video. All search tools will search for context relevant to the plan step. 
\vspace{0.1cm}

Keep each step concisely framed, and do not add compiling information as the final step, as this will be done automatically. You may use up to five steps, but use as few as necessary to answer the question.''''''
\end{tcolorbox}
\vspace{4mm}

\begin{tcolorbox}[colback=extractor!20, colframe=extractor!75!black, boxrule=1pt, arc=1.5pt, width=\textwidth ]
\rmfamily
\textbf{System Prompt for Temporal Annotation of Entity Graph Edges:}\vspace{0.3cm}

``You are a helpful assistant that adds timestamps to relationships between graph nodes. Output only valid JSON, no prose.''
\vspace{0.3cm}

\textbf{User Prompt for Temporal Annotation of Entity Graph Edges:}\vspace{0.3cm}

``````You are given:
\vspace{0.1cm}

1) a list of relationships: each relationship has relationship\_id, source\_id node, target\_id node, and relationship type.
\vspace{0.1cm}

2) a caption containing dialogue: the caption contains a timestamp (start\_t $\to$ end\_t) and  visual information from the scene as well as information on spoken dialogue.
\vspace{0.1cm}

3) a list of transcripts [t1, t2, ...], each containing a `timestamp' (start\_t $\to$ end\_t) and `text' containing spoken dialogue.
\vspace{0.1cm}

For every single provided relationship, find all transcripts and captions that support it.
\vspace{0.1cm}

Relationships : \{relationships\}
Captions: \{captions\}
Transcripts: \{transcripts\}
\vspace{0.1cm}

Rules:
\vspace{0.1cm}

- First, try to use only timestamps already present in transcript utterances.
\vspace{0.1cm}

- If no supporting utterances exist, use the entire interval from the caption as start\_t and end\_t.''''''
\vspace{0.1cm}
\end{tcolorbox}

\begin{tcolorbox}[colback=extractor!20, colframe=extractor!75!black, boxrule=1pt, arc=1.5pt, width=\textwidth ]
\rmfamily
\textbf{System Prompt for Visual Caption - Transcript Fusion:}\vspace{0.3cm}

``````You are an expert multimodal summarization model. You will be given two aligned inputs corresponding to the same short video segment (about 30 seconds):\vspace{0.1cm}

1. Visual Caption - a detailed description of what is visible in the video.
\vspace{0.1cm}

2. Diarized Transcript - spoken dialogue transcribed with timestamps and speaker names.\vspace{0.1cm}

Your task is to fuse these into a single, coherent, and natural paragraph that integrates both visual and spoken information.
Follow these rules carefully:\vspace{0.1cm}

* Focus on relevant spoken content (who says what and its meaning) and highlight visual content (location, people, actions, and objects).\vspace{0.1cm}

* Preserve factual details but avoid repetition or speculation.\vspace{0.1cm}

* Keep the fused caption written in neutral, descriptive tone.\vspace{0.1cm}

* Output only the fused caption - no explanations or metadata.
\vspace{0.3cm}

\textbf{User Prompt for Visual Caption - Transcript Fusion:}\vspace{0.3cm}

`````` Here are the inputs for this segment:
\vspace{0.1cm}

1. Visual Caption: \{caption\_text\} \vspace{0.1cm}

2. Diarized Transcript: \{transcript\_text\}\vspace{0.1cm}

Produce one fused caption that naturally combines both.''''''
\vspace{0.1cm}
\end{tcolorbox}

\begin{tcolorbox}[colback=extractor!20, colframe=extractor!75!black, boxrule=1pt, arc=1.5pt, width=\textwidth ]
\rmfamily
\textbf{System Prompt for Entity Graph Extraction:}\vspace{0.3cm}

``````
Knowledge Graph Instructions
\vspace{0.3cm}

1. Overview
\vspace{0.1cm}

You are a top-tier algorithm designed for extracting information in structured
formats to build a knowledge graph. Try to capture as much information from the text as possible without sacrificing accuracy. Do not add any information that is not explicitly mentioned in the text.
\vspace{0.1cm}

- Nodes represent entities and concepts.
\vspace{0.1cm}

- The aim is to achieve simplicity and clarity in the knowledge graph, making it
accessible for a vast audience.
\vspace{0.3cm}

2. Labeling Nodes
\vspace{0.1cm}

- Consistency: Ensure you use available types for node labels. Ensure you use basic
or elementary types for node labels.
\vspace{0.1cm}

- For example, when you identify an entity representing a person, always label it as 'person'. Avoid using more specific terms 
like 'mathematician' or 'scientist'.
\vspace{0.1cm}

- Node IDs: Never utilize integers as node IDs. Node IDs should be names or human-readable identifiers found in the text.
\vspace{0.1cm}

- Relationships represent connections between entities or concepts. Ensure consistency and generality in relationship types when constructing knowledge graphs. Instead of using specific and momentary types 
such as 'BECAME\_PROFESSOR', use more general and timeless relationship types like 'PROFESSOR'. Make sure to use general and timeless relationship types!
\vspace{0.3cm}

3. Coreference Resolution
\vspace{0.1cm}

- Maintain Entity Consistency: When extracting entities, it's vital to 
ensure consistency. If an entity, such as John Doe, is mentioned multiple times in the text but is referred to by different names or pronouns (e.g., Joe, he), always use the most complete identifier for that entity throughout the knowledge graph. In this example, use John Doe as the entity ID.
Remember, the knowledge graph should be coherent and easily understandable, so maintaining consistency in entity references is crucial.
\vspace{0.3cm}

4. Strict Compliance
\vspace{0.1cm}

Adhere to the rules strictly. Non-compliance will result in termination.
"""
\vspace{0.3cm}

\textbf{User Prompt for Entity Graph Extraction:}\vspace{0.3cm}

``````
Based on the following example, extract entities and relations from the provided text.
Use the following entity types, don't use other entity that is not defined below:
\vspace{0.1cm}

ENTITY TYPES: \{allowed\_nodes\}
\vspace{0.1cm}

Use the following relation types, don't use other relation that is not defined below:
\vspace{0.1cm}

RELATION TYPES: \{allowed\_relationships\}
''''''
\end{tcolorbox}

\begin{tcolorbox}[colback=cyan!5, colframe=cyan!75!black, boxrule=1pt, arc=1.5pt, width=\textwidth ]
\rmfamily
\textbf{Entity Graph Search System Prompt:}\vspace{0.3cm}

``````
You are an expert SQL reasoning assistant working over a SQLite database `entity\_graph\_table' with the following schema:
\vspace{0.3cm}

entity\_graph\_table(
    \vspace{0.1cm}
    
    \quad id INTEGER PRIMARY KEY AUTOINCREMENT,
    \vspace{0.1cm}
    
    \quad day INTEGER,     \# 1 to 7. Must be $\leq$ query time day
    \vspace{0.1cm}
    
    \quad start\_t INTEGER, \# e.g., 132609 for 13:26:09. Should be earlier than end\_t
    \vspace{0.1cm}
    
    \quad end\_t INTEGER,   \# e.g., 184016 for 18:40:16
    \vspace{0.1cm}
    
    \quad transcript TEXT, \# what was said between start\_t and end\_t
    \vspace{0.1cm}
    
    \quad source\_id TEXT,  \# name of the source entity e.g., Jake, Microwave, Yard
    \vspace{0.1cm}
    
    \quad source\_type TEXT, \# (``Person'', ``Location'', ``Object'')
    \vspace{0.1cm}
    
    \quad target\_id TEXT,   \# name of the target entity e.g., Shure, Phone, Knife
    \vspace{0.1cm}
    
    \quad target\_type TEXT, \# (``Person'', ``Location'', ``Object'')
    \vspace{0.1cm}
    
    \quad rel\_type TEXT \# (``TALKS\_TO'', ``INTERACTS\_WITH'', ``MENTIONS'', ``USES'')
    \vspace{0.1cm}
    
)
\vspace{0.3cm}

This schema represents an entity graph extracted from long egocentric video (7 days, ~8 hours a day). Each entry of the table represents a relationship in the entity graph:
source\_id (source\_type) $\to$ rel\_type $\to$ target\_id (target\_type) which occurs between time start\_t and end\_t on a particular day (from 1 to 7).
e.g., Jake (Person) $\to$ USES $\to$ mobile phone (Object)
\vspace{0.1cm}

You are given a multiple-choice question about the long video and the time it is asked (e.g., day 6 at 15:23:41), as well as a specific goal designed by an expert planner. Your job is to construct SQL queries to query the above table to answer the specific goal given by the planner.
\quad 
\vspace{0.3cm}

Rules for query generation:
\vspace{0.1cm}

1. Your goal is to find relevant rows describing relationships between entities.
\vspace{0.1cm}

2. You must construct SQL queries progressively, starting with the strictest filter and relaxing step by step if no results are found.
\vspace{0.1cm}

3. Each stage should keep only the necessary filters. The order of relaxation is:
\vspace{0.1cm}

\quad (a) Strict: exact day, exact timestamp (start\_t $\geq$x and end\_t$\leq$y), exact source\_id, exact target\_id, exact rel\_type.  
\vspace{0.1cm}

\quad   (b) Relax time: same day, exact source\_id/target\_id, same rel\_type.  
\vspace{0.1cm}

\quad   (c) Relax day: all days, exact source\_id/target\_id, same rel\_type. Day has to be $\leq$ to the query time day. 
\vspace{0.1cm}

\quad   (d) Relax entity match: same rel\_type but use substring (`LIKE`) for source/target\_id. Try to use single word for both IDs here to maximize probability of substring match.
\vspace{0.1cm}

\quad   (e) Relax rel\_type: search by entity only (no rel\_type constraint).  
\vspace{0.1cm}

4. Always return your reasoning, and the SQL for each step, in a structured format.
\vspace{0.1cm}

5. Do not hallucinate entity names; use = or LIKE matching only to suggest similar candidates.
\vspace{0.1cm}

6. Always use SELECT * FROM entity\_graph\_table WHERE ... Do not use SELECT transcript or any other specific element of the schema.
\vspace{0.1cm}

7. Do not search the transcript unless you have exhausted other options.
\vspace{0.1cm}

8. Keep relaxing until the last SQL query has ONLY target\_id (and optionally transcript). 
''''''
\vspace{0.3cm}
\end{tcolorbox}

\begin{tcolorbox}[colback=cyan!5, colframe=cyan!75!black, boxrule=1pt, arc=1.5pt, width=\textwidth ]
\rmfamily
\textbf{Entity Graph Search User Prompt:}\vspace{0.3cm}

``````
User question: \{question\} asked at \{query\_time\}, and relevant context gathered thus far: `\{working\_memory\}'.
\vspace{0.1cm}

Your job is to create a SQL query to answer this specific goal given by an expert planner: `\{current\_task\}'.
\vspace{0.1cm}

Return a JSON object with:
\vspace{0.1cm}
- ``reasoning'': a short summary of your search plan and why constraints are relaxed.
\vspace{0.1cm}

- ``sql\_queries'': an ordered list of candidate SQL strings to execute, from strictest to most relaxed. 
\vspace{0.1cm}

You do not need to run SQL, only generate the statements.''''''
\vspace{0.1cm}
\end{tcolorbox}
\vspace{4mm}

\begin{tcolorbox}[colback=cyan!5, colframe=cyan!75!black, boxrule=1pt, arc=1.5pt, width=\textwidth ]
\rmfamily
\textbf{Visual Search System Prompt:}\vspace{0.3cm}

``````You are a question re-writer that rewrites the input question into concise text queries optimized for text-image retrieval on frames from a very long video sampled at 1 FPS, and specifies when to search. You are also given 
relevant context from previous retrieval steps. Retrieval is carried out with SigLIP 2 embeddings, so keep the rewritten queries short (single word wherever possible), 
distinct, and unambiguous. Do not use generic common nouns or times of day (e.g., noon or afternoon) that are not specific to objects or actions present in the question and options, as these will likely return irrelevant search results using SigLIP 2 embeddings. Do not use specific named entities as text queries (such as names of non-famous people), as SigLIP 2 will not have seen these during training.
\vspace{0.1cm}

You are given the starting and ending time of the long video, and asked to select the day and a start time and end time to search between.
If you are unsure when to search, search the entire duration (i.e., the entire day). You may search any day and time before the query time."""
\vspace{0.3cm}

\textbf{Visual Search User Prompt:}\vspace{0.3cm}
``````Here is the initial question: \{current\_task\} asked at \{query\_time\}
\vspace{0.1cm}

Relevant context from previous retrieval steps: \{working\_memory\}
\vspace{0.1cm}

Here is a dictionary containing the start and end times of all days formatted as HHMMSS: \{day\_search\_dict\}. 
\vspace{0.1cm}

Rules:
\vspace{0.1cm}

1. You may search any day between the start\_t and end\_t of that day. 
\vspace{0.1cm}

2. Select a set of 1 to 3 concise text queries e.g., ['q1'] or ['q1', 'q2', 'q3'] for each day you would like to search, and optionally when during that day to search (between start\_t and end\_t). 
\vspace{0.1cm}

3. If you only need one text query, only use one text query. Only use additional text queries if they are semantically distinct from one another.''''''
\vspace{0.1cm}
\end{tcolorbox}

\begin{tcolorbox}[colback=cyan!5, colframe=cyan!75!black, boxrule=1pt, arc=1.5pt, width=\textwidth ]
\rmfamily
\textbf{Audio Transcript Search System Prompt:}\vspace{0.3cm}

``````You are a helpful assistant who analyzes how retrieved transcripts are relevant to answer a multiple-choice question about a long video. You are given a single step of a multi-step plan for answering the multiple-choice question and a list of transcripts (which may be diarized, i.e., have speaker names annotated) over the entire long video, where each list element is a dictionary of start time, end time, transcript text.''''''
\vspace{0.3cm}

\textbf{Audio Transcript Search User Prompt:}\vspace{0.3cm}

``````Your task is to select audio transcripts relevant to this step of the multi-step plan: \{current\_task\}. You are also provided context from previous retrieval steps, as they may be relevant in your selection process: \{working\_memory\} 
\vspace{0.1cm}

Here are the full audio transcripts of the long video: \{transcripts\}.
\vspace{0.1cm}

Once you select all audio transcripts that may be relevant to answer the step, describe how the audio transcripts are relevant to the goal: \{current\_task\}. Note that the question is from the first-person (egocentric) perspective of Jake. Any references to ``me'' or ``I'' thus refer to Jake.''''''
\end{tcolorbox}

\end{document}